\documentclass{article}

 \usepackage[preprint]{neurips_2025}

\usepackage[utf8]{inputenc} 
\usepackage[T1]{fontenc}    
\usepackage{hyperref}       
\usepackage{url}            
\usepackage{booktabs}       
\usepackage{amsfonts}       
\usepackage{nicefrac}       
\usepackage{microtype}      
\usepackage{xcolor}         
\usepackage{graphicx}
\usepackage{amsmath}
\usepackage{subcaption}
\usepackage{enumitem}
\usepackage{todonotes}
\usepackage{colortbl}
\usepackage{placeins}
\title{AirTrafficGen: Configurable Air Traffic Scenario Generation with Large Language Models}

\workshoptitle{LAW 2025: Bridging Language, Agent, and World Models}

\author{%
    Dewi Sid William Gould\\
    The Alan Turing Institute\\
    London, England, NW1 2DB\\
    United Kingdom\\
    \texttt{dgould@turing.ac.uk}\\
    \And
    Benjamin Carvell\\
    NATS\\
    Whiteley, England, PO15 7FL\\
    United Kingdom\\
    \texttt{benjamin.carvell@nats.co.uk}\\
    \And
    George {De Ath}\\
    University of Exeter\\
    Exeter, England, EX4 4QJ\\
    United Kingdom\\
    \texttt{g.de.ath@exeter.ac.uk}\\
    \And
    Nick Pepper\\
    The Alan Turing Institute\\
    London, England, NW1 2DB\\
    United Kingdom\\
    \texttt{npepper@turing.ac.uk}
}

\begin{document}

\maketitle

\begin{abstract}
The manual design of scenarios for Air Traffic Control (ATC) training is a demanding and time-consuming bottleneck that limits the diversity of simulations available to controllers. To address this, we introduce a novel, end-to-end approach, \texttt{AirTrafficGen}, that leverages large language models (LLMs) to automate and control the generation of complex ATC scenarios. Our method uses a purpose-built, graph-based representation to encode sector topology (including airspace geometry, routes, and fixes) into a format LLMs can process. Through rigorous benchmarking, we show that state-of-the-art models like Gemini 2.5 Pro, OpenAI o3, GPT-oss-120b and GPT-5 can generate high-traffic scenarios while maintaining operational realism. Our engineered prompting enables fine-grained control over interaction presence, type, and location. Initial findings suggest these models are also capable of iterative refinement, correcting flawed scenarios based on simple textual feedback. This approach provides a scalable alternative to manual scenario design, addressing the need for a greater volume and variety of ATC training and validation simulations. More broadly, this work showcases the potential of LLMs for complex planning in safety-critical domains.
\end{abstract}

\section{Introduction}

Air traffic control is a complex, safety-critical task that necessitates rigorous selection and training of new air traffic control officers~\citep[ATCOs,][]{cap2331}. Trainee competency is assessed in simulations using handcrafted traffic scenarios designed to test specific skills: for example, recognizing and resolving potential aircraft conflicts. The complexity of these scenarios is maintained at a level appropriate for training. Designing such scenarios is therefore demanding, time-consuming and expensive, requiring significant expert resource. This limitation restricts both the number and the diversity of training scenarios. These challenges apply equally to the construction of scenarios for validating proposed changes to operating procedures, forming a barrier to entry for effective airspace change~\citep{airspace_validation}.

\begin{figure}[htbp] 
    \centering
    \includegraphics[width=1.0\textwidth,trim={0 10 0 0}, clip]{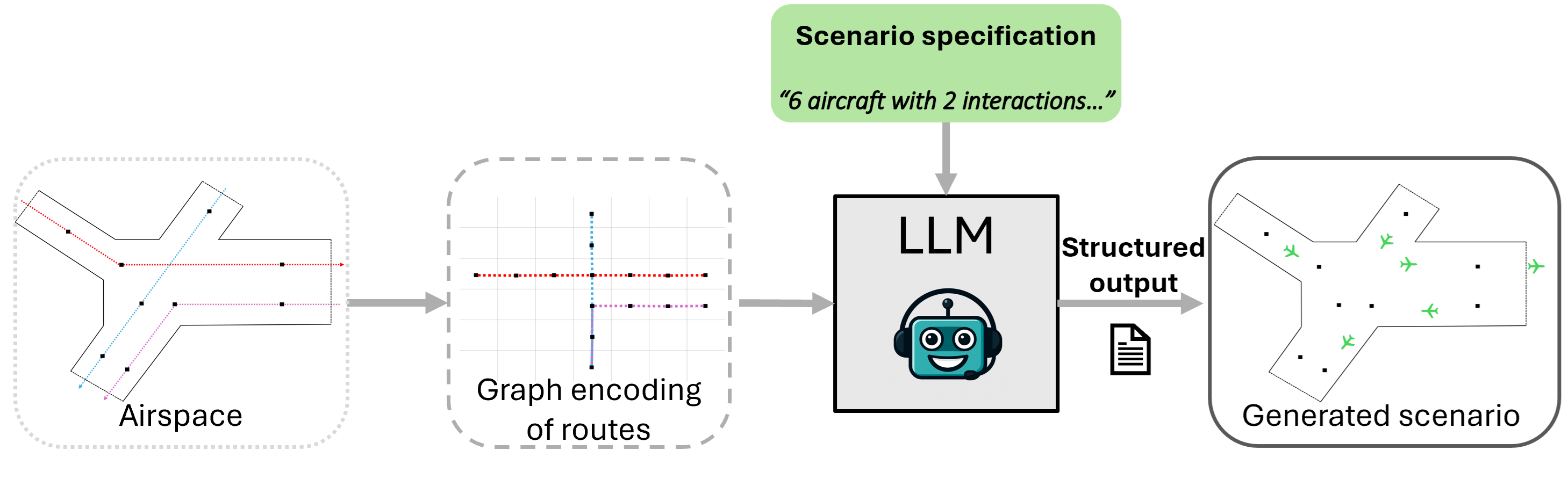} 
    \caption{Overview of the \texttt{AirTrafficGen} framework for \textbf{fully configurable} scenario generation on \textbf{arbitrary} airspace geometries. Given a sector of airspace (a three-dimensional volume with prescribed routes which aircraft can fly on - denoted by coloured lines) and a \textbf{text-based scenario specification}, we engineer a prompt which encodes the relevant spatial information for an LLM to create a highly specific air traffic scenario. The framework converts the sector geometry into a grid-based graph, designed using key air traffic control safety length-scales. The output is in a structured JSON format, which we feed in to a simulation environment.}
    \label{fig:paper_overview}
\end{figure}

Controlled airspace is divided into geographical units known as sectors, each governed by sector-specific procedures. Hence, designing representative validation scenarios requires substantial domain expertise and sector-specific knowledge. Automating scenario generation in both contexts could markedly increase scenario quantity and diversity. Therefore, a method is needed to \emph{controllably} generate new interacting scenarios while respecting existing route structures and sector procedures. This paper explores the novel application of Large Language Models~\citep[LLMs,][]{DBLP,radford2018improving} as a principled alternative to handcrafted scenario generation.

We present an end-to-end framework for fully configurable scenario generation: \texttt{AirTrafficGen}. A core novelty is our rigorous benchmarking of LLM capabilities on this complex task. First, we introduce a novel mapping from three-dimensional airspace sectors to discrete graphs, engineered to fit within an LLM's context window. Within this setting, we systematically benchmark the reasoning capabilities of state-of-the-art LLMs across orthogonal subtasks required for scenario generation. Specifically, we evaluate competency along four reasoning axes: (1) \textbf{aircraft spatial density} -- handling higher traffic loads; (2) \textbf{temporal reasoning} -- varying scenario duration; (3) \textbf{sector complexity} -- adjusting route interactivity; and (4) \textbf{interaction modelling} -- engineering scenarios of differing complexity. Our benchmarks test LLM ability to generate scenarios with differing aircraft counts, durations, and sector complexities, providing granular insight into model strengths and limitations. Finally, we implement an end-to-end pipeline that demonstrates full controllability and practical utility in producing realistic, challenging air traffic scenarios. See Figure~\ref{fig:paper_overview} for a concise overview.

The contributions of our paper are as follows:
\begin{itemize}
    \item Novel benchmarking showcasing the strengths and limitations of LLMs in \textbf{spatial}, \textbf{temporal}, and \textbf{spatio-temporal} reasoning crucial for complex air traffic scenario generation.
    \item A purpose-built, \textbf{graph-based knowledge representation} that enables LLMs to ingest and reason over intricate spatio-temporal air traffic data.
    \item An \textbf{engineered-prompting framework} that provides fine-grained control over scenario characteristics including interaction presence, type and location.
    \item Empirical demonstration of the method's applicability across diverse route configurations.
\end{itemize}

\section{Background on Air Traffic Control}
\begin{figure}[htbp]
    \centering
    \begin{subfigure}[t]{0.59\linewidth}
        \centering
        \includegraphics[width=\linewidth]{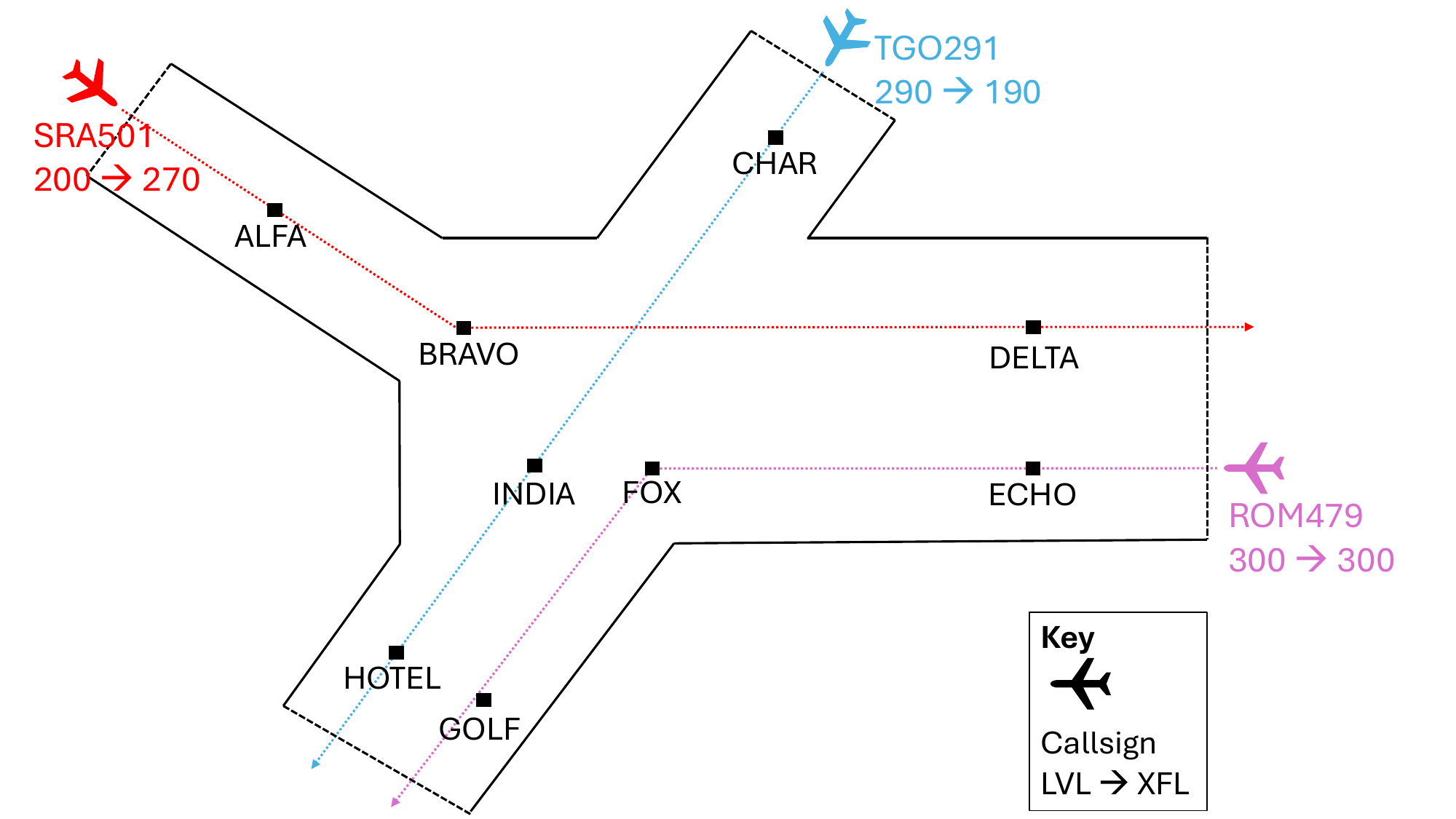}
        \caption{A schematic illustrating an airspace. Fixes are denoted with black boxes, and routes flown by aircraft are marked with coloured lines.}
        \label{fig:relevant_aircraft}
    \end{subfigure}%
    \hfill
    \begin{subfigure}[t]{0.39\linewidth}
        \centering
        \includegraphics[width=\linewidth]{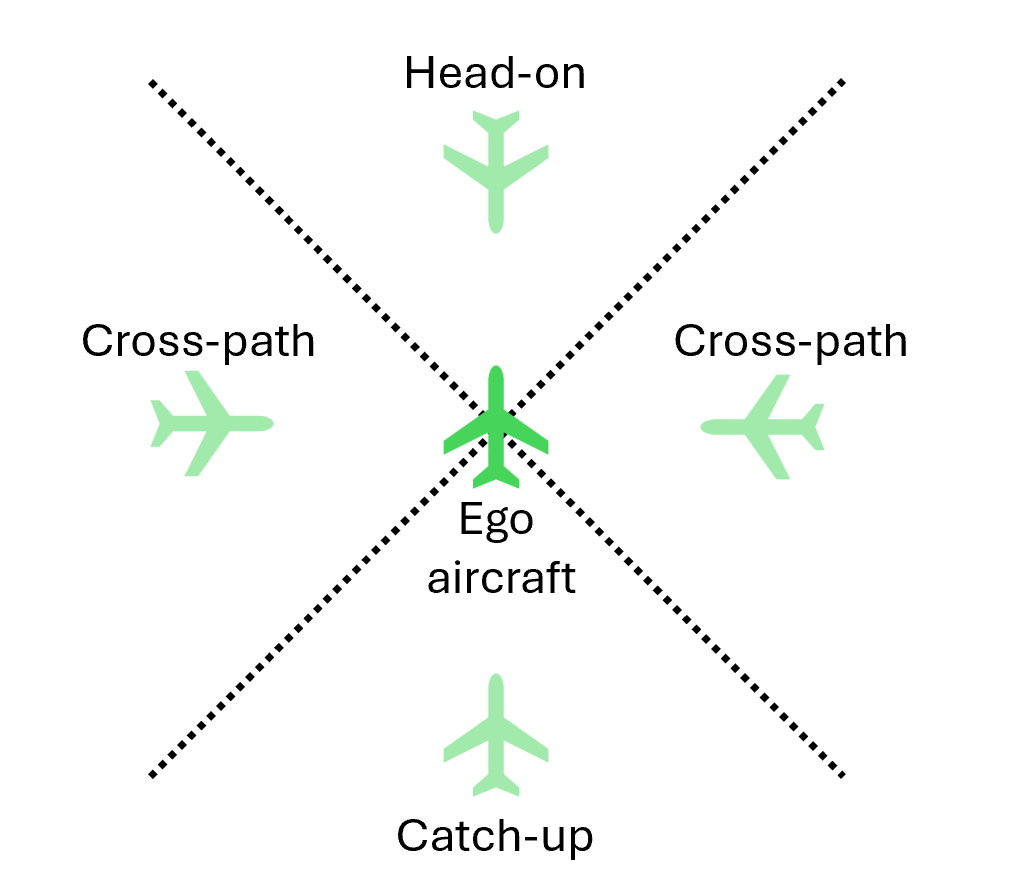}
        \caption{The three fundamental pairwise aircraft interaction types: cross-path, head-on and catch-up.}
        \label{fig:interaction-definitions}
    \end{subfigure}
    \caption{Overview of relevant airspace (left) and aircraft interaction types (right).}
    \label{fig:airspace_and_interactions}
\end{figure}

The role of an ATCO is to ensure the safe, orderly and expeditious transit of aircraft through their sector. Before each flight, the aircraft operator submits a flight plan specifying the route as a sequence of GPS waypoints, or \textit{fixes}. This plan informs the sector ATCO of the aircraft's intended lateral track and its requested \textit{exit flight level} (the altitude at which it will leave the sector). Flight levels are reported in hundreds of feet (e.g., FL~$250$ corresponds to $25,000$ft). Figure~\ref{fig:relevant_aircraft} depicts a sector and its fixes. Because the sector contains only a finite set of fixes, the number of possible routes through it is also finite. The $i$\textsuperscript{th} aircraft within a scenario can be characterised by the following information: aircraft-type, spawn time, initial flight-level, $h_i$, requested/exit flight-level, $e_i$, and route.



An ATCO's overriding priority is to maintain separation minima between all aircraft in the sector. They must formulate deconfliction plans that remain robust even in the event of communication failure, greatly increasing task complexity. The task is further complicated by the presence of large epistemic uncertainties concerning aircraft performance (see, e.g., \citealp{pepper_jais}). Aircraft pairs are deemed to be \emph{relevant traffic} when the ATCO may need to issue instruction to assure safety, at ranges that are much greater than the separation standards for that airspace. No formal metric determines if two aircraft are relevant to one another; this depends on the ATCO's judgement, the specific sector, and the wider operational context. For this reason, we define a \textit{relevancy metric}, based on discussions with ATCOs. Under this metric, all aircraft in a scenario are non-interacting if, on their current trajectories, no pair with overlapping flight level ranges\footnote{The \textit{flight level range} of an aircraft is defined as the range $(\text{min}(h_i, e_i), \text{max}(h_i, e_i))$.} come within $20$ nautical miles (nmi) of one another. This threshold equates to a 2--3 minute look-ahead time depending on the ground speed of the aircraft, which is comparable to the timescales used in short-term conflict detection (see, e.g. \citealt{RADANOVIC2018105}).\looseness=-1

A \emph{non-interacting} scenario is one where no aircraft are considered relevant to each other, allowing them to operate independently. Independent aircraft reduce complexity because instructions issued to one need not consider the safety of others. A constraint when designing air traffic scenarios is that an aircraft should not be relevant traffic for at least $2$ minutes after entering the controlled sector. This reflects operations within a sectorised airspace, in which aircraft are passed between different sectors in such a way that they do not pose an immediate safety issue in the new sector~\citep{coordination}.

\emph{Interactive} scenarios exhibit several distinguishing features. At a high level, specifying the number of interactions, and \textit{types} of these interactions, is an effective measure of ``interactivity" or scenario \textit{complexity}. Interactions are classified into three types according to the configurations of the participating aircraft (see Figure~\ref{fig:interaction-definitions}). At the level of classification, any multi-aircraft interaction can be decomposed into groups of pairwise interactions.\looseness=-1

\section{Related Work}
Earlier studies generated air traffic scenarios by algorithmically re-working recorded data to insert interactions~\citep{scen_gen_alg2,scen_gen_alg} or by keeping a human in the loop~\citep{scen_gen_human}. More recently, \citet{Stefani2025Automated} proposed an automated generator to validate machine learning conflict-resolution tools. Yet few studies create realistic air traffic scenarios from scratch in synthetic airspace.

In contrast, machine learning methods have been used to generate complex scenarios across domains as varied as economics~\citep{economics}, healthcare~\citep{health}, energy~\citep{energy}, and transportation~\citep{trafficgen}. In transportation, scenario-based testing is used to assess autonomous driving systems \cite{self_driving_cars, AV_scengen_review}. Several advanced, data-intensive methods such as diffusion~\citep{diffscene} and generative adversarial networks~\citep[GANs,][]{AV_GANS} have been investigated to generate automotive scenarios. However, these data-intensive methods are poorly suited to our setting, where the synthetic airspace datasets for ATCO trainees are small.

Road-traffic scenarios resemble air traffic ones: both involve multiple vehicles interacting on procedurally constrained route structures. Road traffic is tightly constrained by lanes and rules, whereas aircraft have three-dimensional freedom as ATCOs can direct them off their filed routes. Safety definitions also differ: air traffic management employs high redundancy and conservative thresholds, making fine-grained controllability harder. Moreover, road-traffic scenarios typically revolve around a \textit{single ``ego-vehicle"}, whereas air traffic scenarios evaluate a controller's handling of the \textit{complete traffic flow}. Furthermore, road traffic scenarios are primarily benchmarked by collision rates involving an agent-controlled ego vehicle~\citep{safebench}. In contrast, air traffic scenarios lack such an agent, and safety is measured in a more subtle manner: the controlling technique of ATCOs is assessed across a diverse range of competencies based on different configurations of aircraft. This difference requires designing ATC scenarios with a unique degree of controllability over scenario characteristics.

A key requirement for testing automotive driving agents is the ability to generate automotive scenarios controllably, allowing the user to select scenarios from the same distribution as the training data or to adversarially generate scenarios. Hence, LLMs have recently been applied to the generation of road traffic scenarios~\citep{chatscene, llmscenario,lu2024multimodal,text2scenario}. LLM-based generation therefore requires an interpretable encoding of the road network topology. In this paper, geometric information concerning sector routes is encoded in a graph-based representation, which is described in the next section.

\section{Graph Representation of Air Traffic Control Scenarios}
\label{sec:graphrep}
Continuous airspace and aircraft trajectories must be discretised for an LLM framework. The structure of our proposed discretisation is motivated by the length-scale used to determine whether aircraft are relevant traffic to one another. 

\begin{figure}[h]
    \centering
    \begin{subfigure}[b]{0.45\textwidth} 
        \centering
        \includegraphics[width=\textwidth]{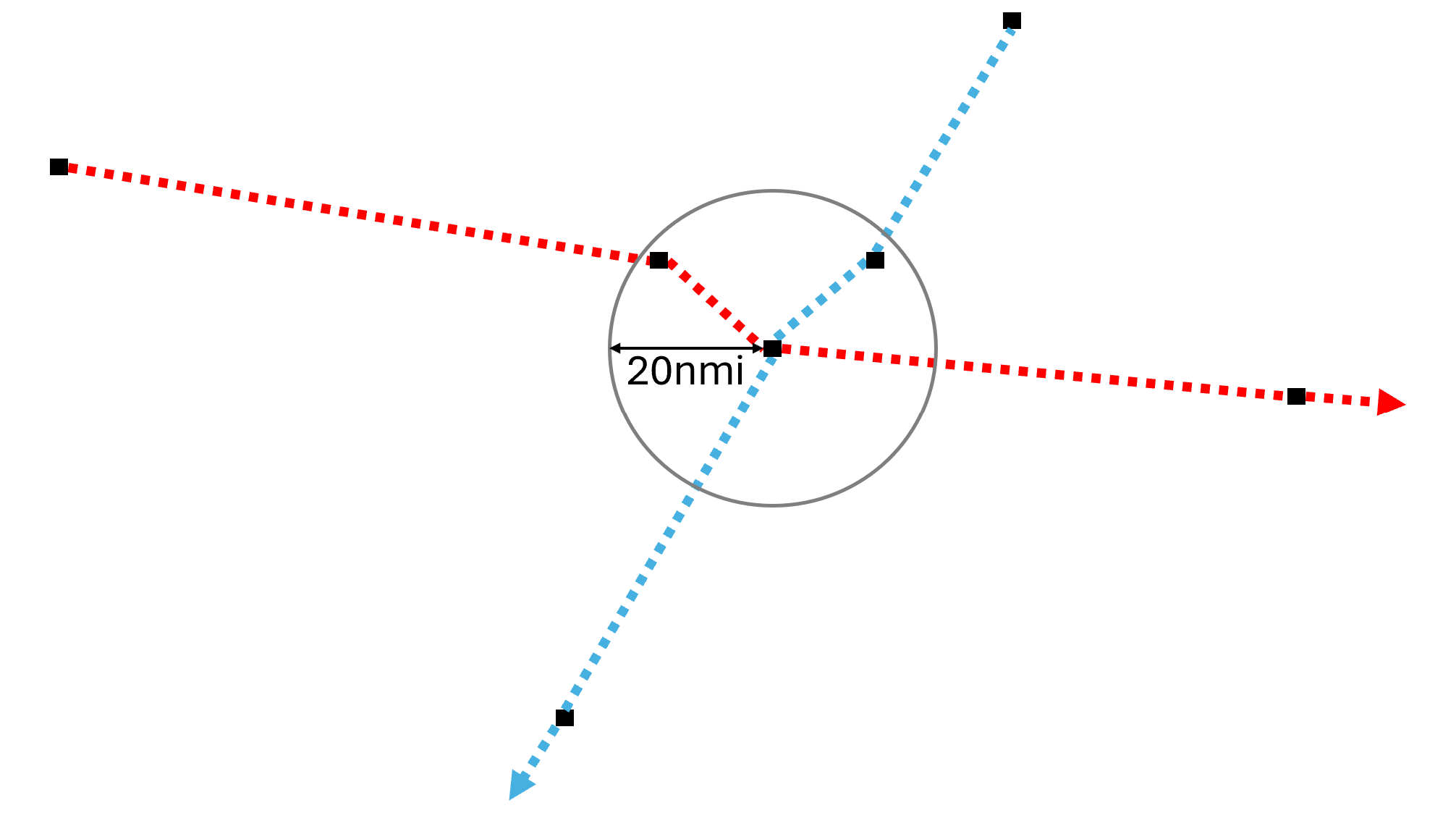}
        \caption{Two example routes prior to graph encoding.}
        \label{fig:graph_encode1}
    \end{subfigure}
    \hfill 
    \begin{subfigure}[b]{0.45\textwidth}
        \centering
        \includegraphics[width=\textwidth, height=0.7\textwidth, trim= 0 2cm 0 1cm, clip]{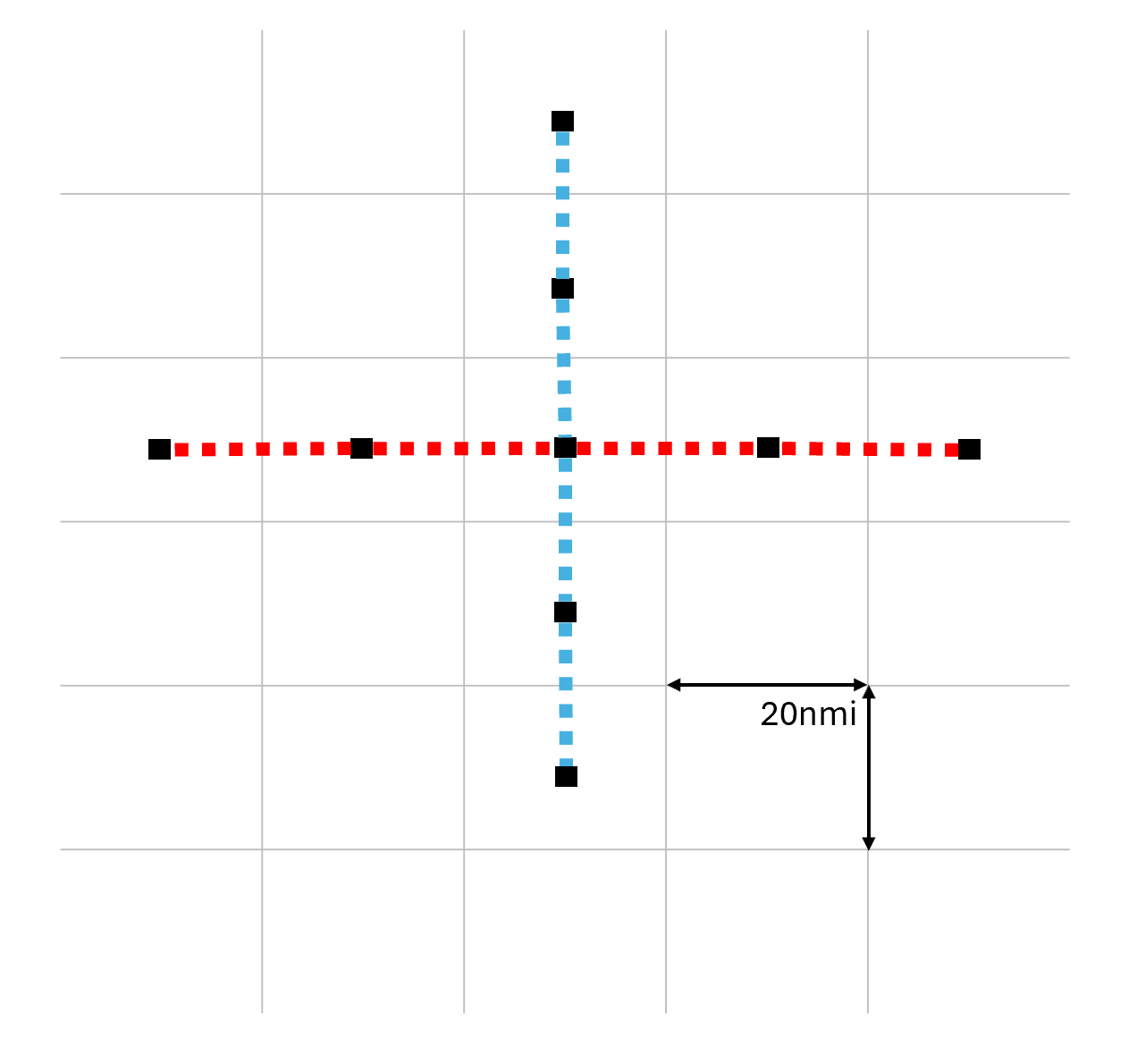}
        \caption{The graph representation of the routes in \ref{fig:graph_encode1}.}
        \label{fig:graph_encode2}
    \end{subfigure}
    \caption{Converting sector routes to a graph representation. The three fixes encircled are all within $20$ nmi, meaning that in the graph representation they will be projected onto the same node. The legs of the routes are interpolated with nodes every 20 nautical miles. Notice that we throw away unimportant kinks in the routes for maximal simplicity, retaining only the route lengths and intersections.}
    \label{fig:routes_graph}
\end{figure}

To represent the spatial relationships between the fixes along routes efficiently for LLM consumption, we project the routes into a graph-based representation, where nodes of the graph are separated by $20\,$nmi. Figure~\ref{fig:routes_graph} illustrates this process for two example routes, with Figure~\ref{fig:graph_encode1} displaying example routes prior to the encoding and Figure~\ref{fig:graph_encode2} the corresponding graph. The process can aggregate multiple fixes into a single graph node, which is a useful feature as sectors typically contain a high density of fixes in proximity.

This method is motivated by the need to retain only essential information for configurable generation: route \textit{length} (to simulate traversal time) and route \textit{intersections} (to manage aircraft interactions). See~\citep{kaggle-game-arena-2025} for some related work involving the encoding of spatial environments for LLMs within the context of games.

\subsection{Aircraft Dynamics}
\begin{figure}[htbp]
    \centering
        \includegraphics[width=1.0\linewidth,trim={0 0 0 0}, clip]{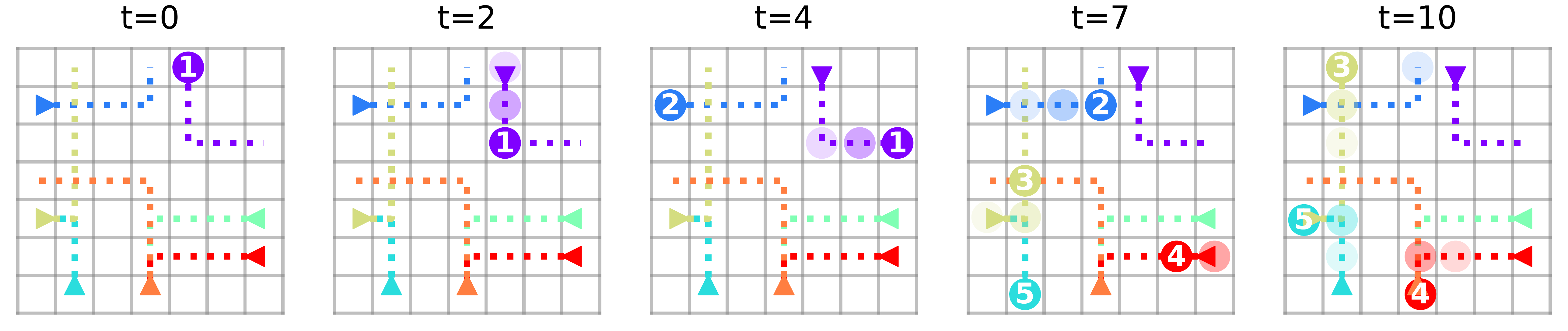}


    \caption{An example \textit{non-interacting} scenario for the traffic volume benchmark. Aircraft spawn at the start of their routes, and move in discretised time units along the route, one grid cell at a time.}
    \label{fig:scenario_and_sector}
\end{figure}

Aircraft traverse the graph along their assigned routes. Each aircraft is classified as fast (one node at each time-step) or slow (one node every two time-steps) to reflect a wide range of performance characteristics present within controlled airspace~\citep{hodgkin2025probabilisticsimulationaircraftdescent}. Slow speeds correspond to turboprops, and fast speeds to jet engines when converting our discretised scheme to any simulator. Due to the large discretisation length-scale and broad definition of relevant traffic, this simplification is designed to have a limited impact on the fidelity of our method. See Figure~\ref{fig:scenario_and_sector} for an example scenario, with Figure~\ref{fig:synthetic_sector} showing the row structure of the synthetic sector in greater detail. 

\subsection{Interactions on the Graph}
In the discrete model, an interaction occurs when (1) two aircraft occupy the same node simultaneously, or (2) two aircraft swap nodes in one time-step. By design, the graph construction ensures that no edges cross without meeting at a node; therefore, these two events capture \emph{every} instance where aircraft pass within the $20$ nmi threshold.



\section{LLM Prompting Framework}
This section details our multistep prompting framework. The full prompts used in this paper are detailed in the Supplementary Material. The LLM receives two inputs: 1) Specification: a text description of the required scenario, and 2) Sector geometry: permitted routes and their layout.
Routes are converted to the graph formalism of Section~\ref{sec:graphrep}.

The prompt first establishes the core task, defining the scenario rules and settings, and provides detailed instructions on how aircraft move and the definition of an interaction. This is followed by a ``high-level strategy'' section which acts as the core of the prompt. This explicitly directs the model to split the task into three phases: 1) Sector analysis (identify route intersections), 2) Aircraft placement strategy, and 3) Internal verification and iterative refinement. 

The framework's key feature is the third phase, where the model is prompted to internally verify its own scenario by rolling out trajectories to check for unintended interactions. We prompt the LLM to return a scenario in a fixed JSON format where each entry includes a spawn time (integer), a route (string identifier), and aircraft speed (1 or 2). The JSON files generated by the LLMs were formatted so that they could be parsed by BluebirdDT, a probabilistic digital twin of en route airspace~\citep{bluebird}. All scenario screenshots used in this work are generated using this twin.\looseness=-1



\section{Benchmarking Scenario Reasoning}
In this section, we evaluate how well contemporary large language models perform each component of air traffic scenario generation. The task provides a natural testing ground for spatial reasoning and planning. To quantify these skills, we introduce four novel benchmarks.

To emulate the manual design process, where lateral conflicts are typically planned before vertical separation is assigned, our benchmarks are initially restricted to the 2D \textbf{lateral plane} (${h_i \equiv e_i \equiv H\,, \forall i}$). This isolates the problem of creating or avoiding lateral conflicts before Section~\ref{sec:controllability_section} introduces the vertical dimension. Handling large aircraft counts or dense interaction patterns demands strong spatial awareness, temporal reasoning and careful forward planning. Succeeding in this task requires simulating how aircraft move, tracking all possible interactions, and making tactical decisions about parameters to best avoid/create them.

\subsection{Benchmark Suite}
We begin with the fundamental spatial and temporal reasoning required to understand air traffic scenarios. Each benchmark asks the LLM to produce a \textit{non-interacting} scenario that meets specific constraints. See Table~\ref{tab:models_benchmarked} for the full list of models we benchmark in this work. To ensure statistical confidence in our results, we test every parameter set on \textbf{ten} synthetic sectors (Figure~\ref{fig:scenario_and_sector} shows one example). The code required to run these benchmarks, including these synthetic sectors, is included in the Supplementary Material. All benchmarks are \textbf{automatically verifiable}: after generating trajectories on the graph, we can directly count the number of interactions.


\paragraph{Traffic Volume Benchmark.} For a fixed scenario length and sector complexity, we measure LLM performance at creating non-interacting scenarios with increasing numbers of aircraft. See Table \ref{tab:num_ac_benchmark_table} for the results of this benchmark.
\begin{equation}
     N \in \{2,3,4,5,6,7,8,9,10,15,20,25,30\}\,, \quad T=12\,, \quad N_{\text{routes}}=7\,, \quad N_{\text{intersections}}=7\,.
\end{equation}
Here, $N$ is the number of aircraft, $T$ is the number of scenario time units, $N_{\text{routes}}$ is the number of routes in the sector and $N_{\text{intersections}}$ is the number of graph nodes occupied by more than one route (a proxy for sector complexity). Figure~\ref{fig:scenario_and_sector} illustrates an example sector used in this benchmark.

\paragraph{Scenario Length Benchmark.} For a fixed number of aircraft and sector complexity, we measure LLM performance at creating non-interacting scenarios of increasing length. See Table \ref{tab:scenario_len_benchmark_table} for the results of this benchmark.
\begin{equation}
    T \in \{12,15,18,21,24 \} \,, \quad N=8\,, \quad N_{\text{routes}}=7 \,, \quad N_{\text{intersections}} = 7 \,.
\end{equation}

\paragraph{Sector Complexity Benchmark.} Next, we design a benchmark to focus on measuring the capability of models to handle sectors of increasing intrinsic complexity (measured by the number of intersection points of the routes in the sector). We do this for a fixed number of aircraft and scenario length. See Table \ref{tab:sector_complexity_table} for the results of this benchmark.
\begin{equation}
\begin{aligned}
    N_{\text{intersections}} \in [4,14] \,, \quad N=8\,, \quad T=12 \,, \quad N_{\text{routes}}=7\,.
\end{aligned}
\end{equation}
Success is measured by the \textbf{mean number of unique pairs of interacting aircraft} (MUIP) across the 10 synthetic sectors; the ideal value is zero. All models are compared with a random baseline score which is computed by averaging over 500 scenarios created by sampling spawn times, routes and speeds from their valid sets uniformly. In Figures~\ref{fig:good_example}~and~\ref{fig:bad_example}, we give two examples of Gemini-2.5-Pro's responses on the $N=30$ traffic volume benchmark, on two different sectors.

Lastly, we design a \textbf{controllability} benchmark, which assesses an LLM's ability to construct a target number of interactions. We use the parameters
\begin{equation}
    T =12 \,, \quad N=10\,, \quad N_{\text{routes}}=7 \,, \quad N_{\text{intersections}} = 7 \,,
\end{equation}
and prompt for $\{1,2,3,4,5\}$ \textit{unique} interacting pairs. Success is gauged by the \textbf{mean absolute difference between number of unique interacting pairs and the input number of interacting pairs} (MADIP)  across the 10 synthetic sectors. See Table \ref{tab:input_conflict_table} and Figure~\ref{fig:num_conflict_box_plot} for the results of this benchmark.

All experiments used the OpenRouter inference platform~\cite{OpenRouter}. Specific prompts for each benchmark are given in the Supplementary Material. For all experiments, we used a temperature of $1.0$, $\text{top\_p} = 1.0$, $\text{top\_k} = 0.0$, and a maximum token budget of $35,000$. For each task, we use the first scenario generated by the LLM that satisfies the required format, rather than generating multiple candidates. If a model failed to produce a valid output due to token limits, the token budget was increased in steps of $10,000$ until success. No model required more than $50,000$ tokens to produce a valid response on all benchmarks. The combined cost of all the experiments carried out in this paper was under $\$150~\text{USD}$.

\subsection{Benchmarking Results}
\begin{figure}
    \centering
    \includegraphics[width=\linewidth, trim={0 0 0 0}, clip]{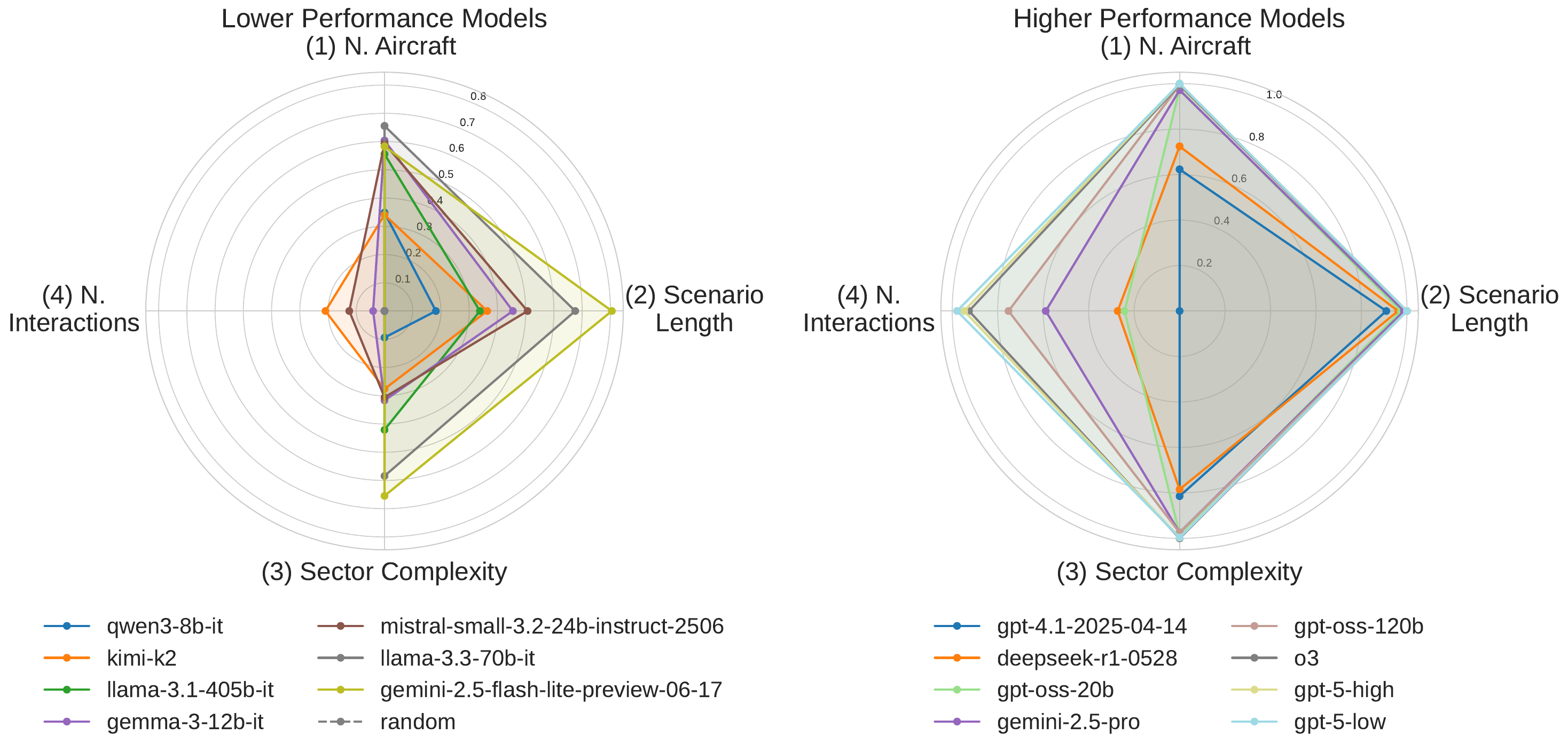}%
    \caption{LLM capabilities across four benchmarked axes.}
    \label{fig:capability_plot}
\end{figure}

The complete benchmark results appear in Appendix~\ref{sec:benchmark_results}. Figure~\ref{fig:capability_plot} summarises model capabilities; here \textit{ability} is measured as a normalised score $\in [0,1]$ relative to the mean score of the random baseline (labelled with the \textit{rand} subscript). The scores are calculated as follows:\looseness=-1
\begin{equation}
    \mu_{1,2,3} = \left[1-\frac{\text{MUIP}}{\text{MUIP}}_{\text{rand}}\right]_+ \quad \mu_4 = \left[1-\frac{\text{MADIP}}{\text{MADIP}}_{\text{rand}}\right]_+,
\end{equation}
where $[x]_+ = \max(0, x)$, a value of $0$ matches or underperforms the random baseline, and $1$ corresponds to perfect performance.

We observe considerable variation in model performance. Interestingly, performance across capability dimensions is not fully correlated. For example, mistral-small-24b-it handles scenario length better than llama-405b-it, but not sector complexity. The most advanced models, GPT-5, o3 and Gemini-2.5-Pro, are the most effective, with GPT-5 performing near-optimally across all benchmarks, whereas Gemini-2.5-Pro falters at the upper end of the interaction task. We also observe strong performance of the open-source GPT-oss models, in particular with the 120b variant outperforming most advanced proprietary models. This result is clearly demonstrated in Appendix \ref{sec:benchmark_results}, Figure \ref{fig:pareto_plot} where we compare model performance against cost - plotting the Pareto frontier.

We observe that most models struggle to perform better than random at the fourth benchmark. It should be noted that the latter elements of the controllability benchmark, involving generating 4 or 5 interactions in a scenario with 10 aircraft, are incredibly difficult and go beyond the scope of design control required to make realistic air traffic scenarios. Moreover, the full task includes the vertical dimension, giving designers an extra degree of freedom to alleviate congestion.

\section{Configurable and Flexible Air Traffic Scenario Generation}
\label{sec:controllability_section}

We have observed that some advanced reasoning models are capable of designing high traffic non-interacting scenarios and produce promising results when prompted to generate interactions in the lateral plane. We therefore turn to the central requirement of realistic simulation: \textbf{fine-grained controllability}. To meet the full complexity and fidelity demanded by operational air traffic control simulations, we add the vertical dimension: each aircraft now carries an initial flight level and a requested exit flight level in the structured JSON output. Human controllers find the task markedly easier once the vertical dimension is available, because potential lateral conflicts can be resolved by altitude separation.

This section demonstrates the breadth of controllable parameters that scenario designers can exploit when generating air traffic scenarios. For illustration, we employ Gemini-2.5-Pro to generate a wide spectrum of specific scenarios. Our testing includes fine-grained instructions with detailed requirements including number of aircraft, conflicts (including location, time and type), as well as general traffic patterns (including features like ``in trail" or ``climbing through levels"). A further strength is the ability to \textbf{adapt existing scenarios} to new requirements. This includes adding new aircraft or modifying the parameters of existing aircraft to precisely tailor a scenario to specific requirements. This adaptability highlights the flexibility and operational relevance of \texttt{AirTrafficGen}.

\begin{figure}
    \centering
    \begin{subfigure}{0.32\textwidth} 
        \centering
        \includegraphics[width=\linewidth]{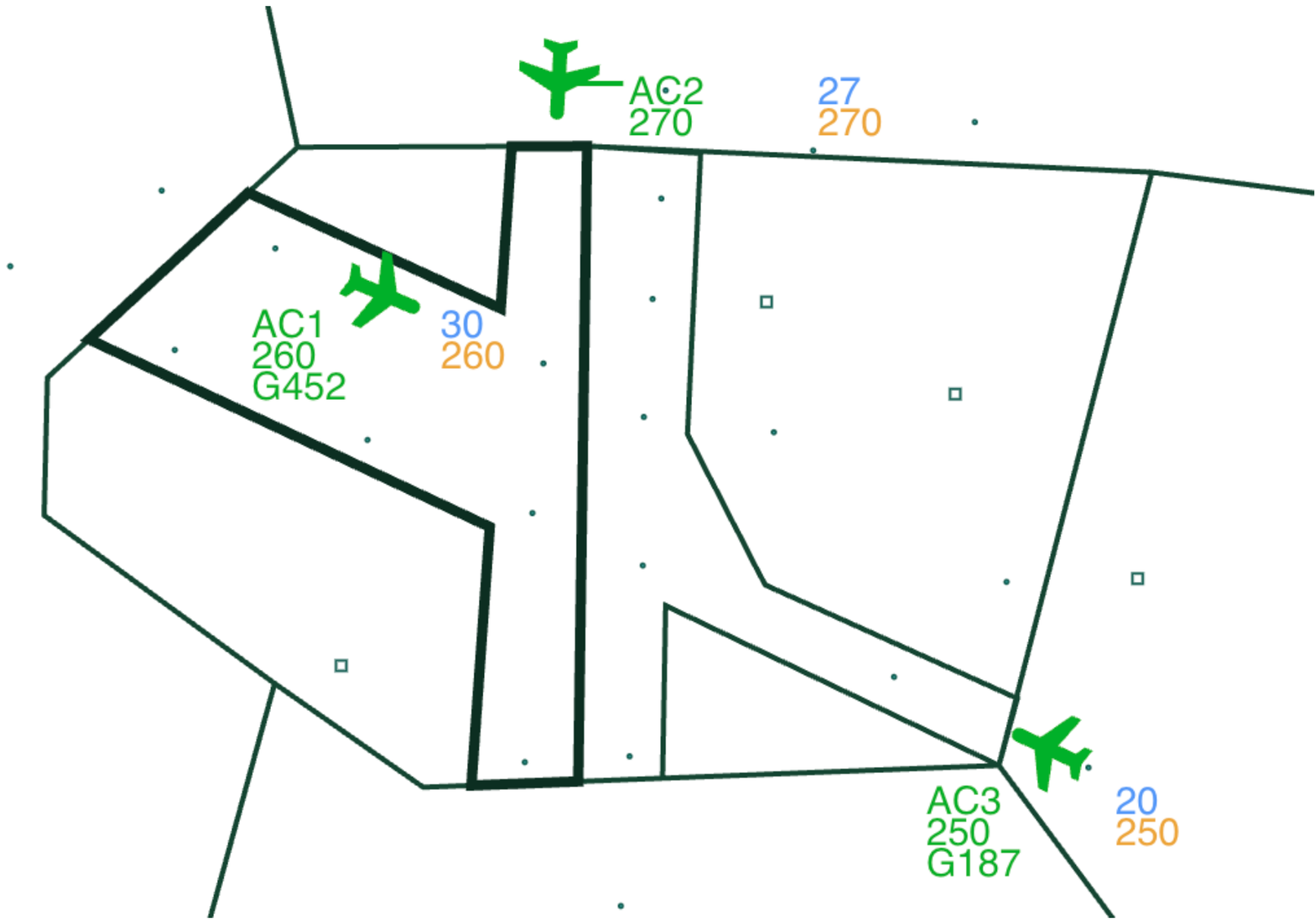}%
        \caption{Generate a scenario with four aircraft. There should be two aircraft which interact in a cross-path manner. The remaining two aircraft should not interact with anything.\protect\\\strut}
        \label{fig:image2}
    \end{subfigure}
    \hfill 
    \begin{subfigure}{0.32\textwidth} 
        \centering
        \includegraphics[width=\linewidth]{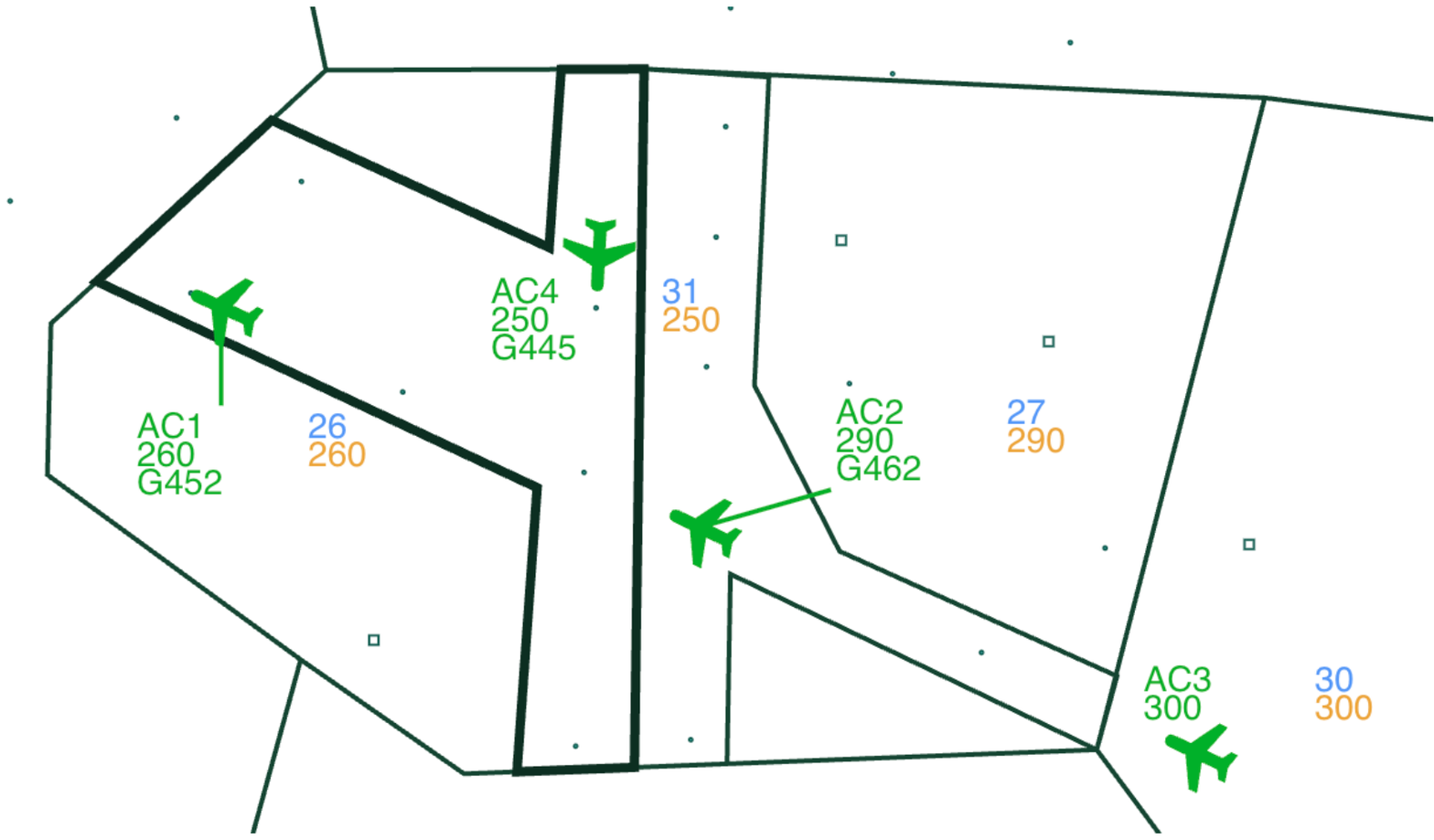}%
        \caption{Generate a scenario with four aircraft. Three aircraft should follow one another in trail, but not interacting. The fourth should be a climber which climbs through all the levels of the three in trail.}
        \label{fig:image4}
    \end{subfigure}
    \hfill 
    \begin{subfigure}{0.32\textwidth} 
        \centering
        \includegraphics[width=\linewidth, trim=0cm 0 0 0 0 clip]{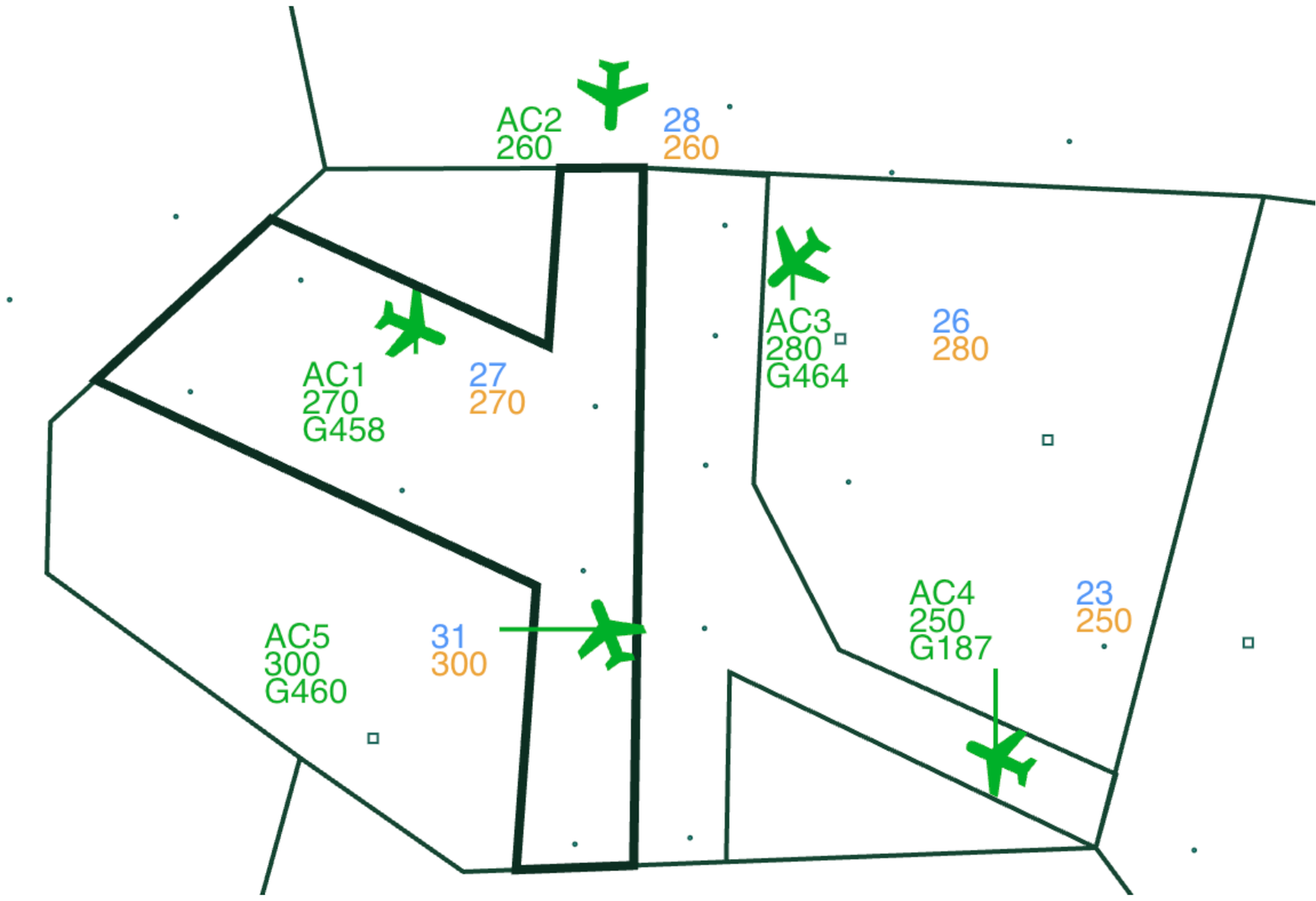}%
        \caption{Generate a scenario with six aircraft. There should be an interaction in which three aircraft are involved. All other aircraft should be non-interacting.\protect\\\strut}
        \label{fig:image6}
    \end{subfigure}
    \caption{Three examples of fine-grained scenario controllability using AirTrafficGen. Each caption details the prompt used. Images are generated using BluebirdDT. Blue numbers are exit flight levels, and orange numbers are initial flight-levels. GXXX represents the ground speed of the aircraft in knots. In (a) the generated interaction is between AC1 and AC2, while the fourth aircraft is yet to spawn. In (b) AC1, AC2 and AC3 are following each other in trail, while AC4 is climbing across their route and through their levels as requested. In (c) the triple interaction is created using AC1, AC2 and AC3.}
    \label{fig:six_images}
\end{figure}

\begin{figure}[htbp]
    \centering
    \begin{subfigure}{0.45\textwidth}
        \centering
        \includegraphics[width=\linewidth]{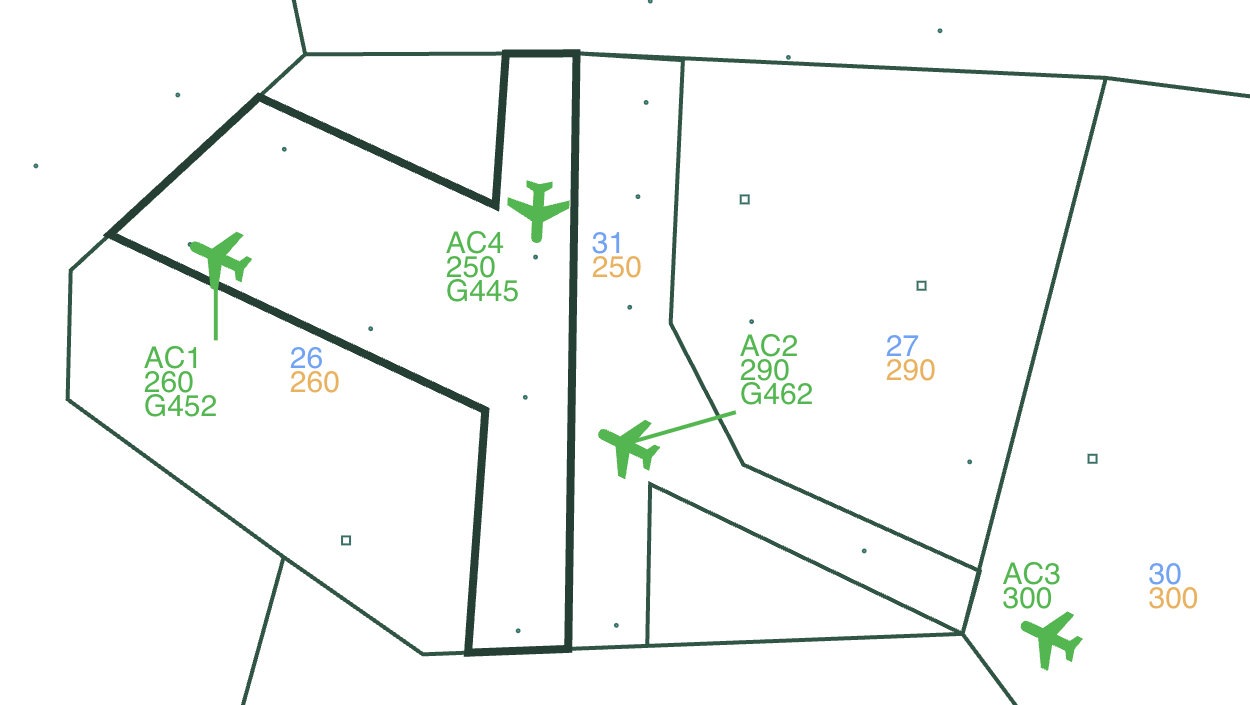}
        \caption{Original scenario.\protect\\\strut}
        \label{fig:your_first_image}
    \end{subfigure}
    \hfill
    \begin{subfigure}{0.45\textwidth}
        \centering
        \includegraphics[width=\linewidth]{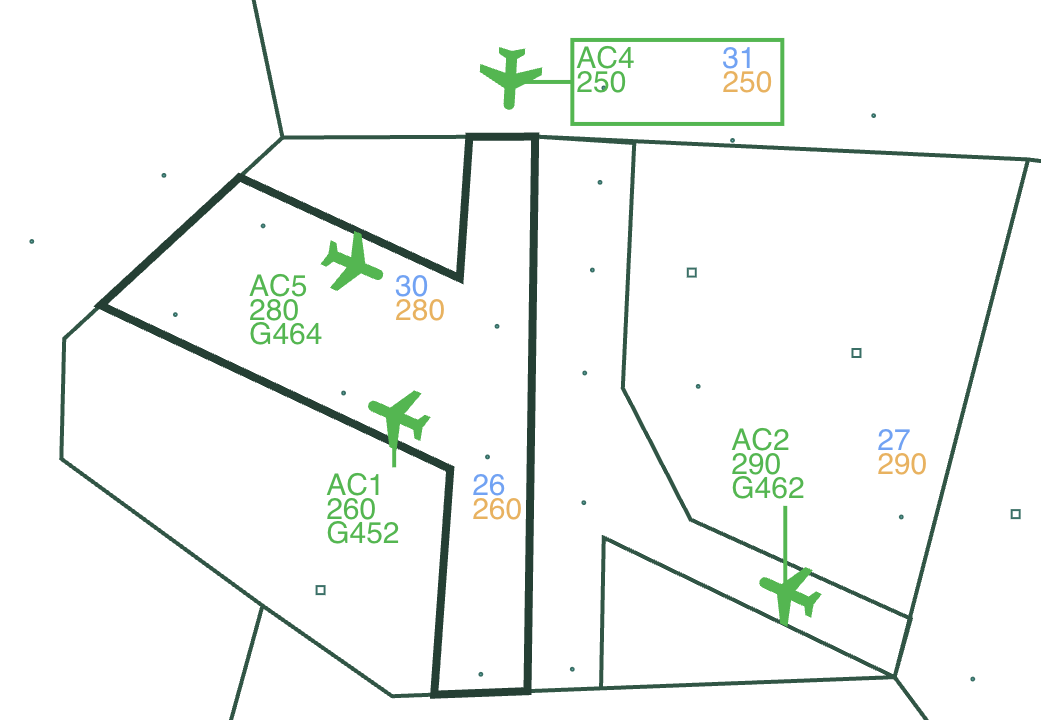}
        \caption{LLM-adapted scenario to make interaction harder.}
        \label{fig:your_second_image}
    \end{subfigure}

    \caption{When prompted to make the interaction in (a) harder to solve by adding a new aircraft, the LLM has added AC5. AC5 has flight level overlap with AC2 and AC4, adding to the complexity of the scenario. An ATCO may solve (a) by climbing AC4 to an intermediary (but still safe) level of $280$. With AC5, this level is now occupied - blocking this possible solution.}
    \label{fig:two_images}
\end{figure}

To illustrate the control offered by \texttt{AirTrafficGen}, we provide concrete examples:
\begin{enumerate}
    \item \textbf{Pairwise Interactions:} 
    Appendix~\ref{sec:controllability_appendix} and Figure~\ref{fig:pairwise_examples} show precisely engineered scenarios with all three fundamental pairwise interactions: cross-path, head-on, and catch-up.
    \item \textbf{Sophisticated Controllability:} Figure~\ref{fig:six_images} presents three complex scenarios that involve a higher number of aircraft, intricate and detailed instructions, and combinations of various control parameters.
    \item \textbf{Scenario Modification:} Finally, in Figure~\ref{fig:two_images}, we provide an example of controllability in terms of \textit{modifying an existing scenario}. This demonstrates how our method can take a pre-existing traffic scenario and adapt it to new requirements, such as increasing its complexity or introducing new elements.
\end{enumerate}

The method is both powerful and generalisable across sectors and operational contexts, and it directly addresses the laborious and time-intensive nature of handcrafting scenarios. Although the full complexity of high-traffic scenarios is challenging to convey in static images, the provided screenshots with lower traffic densities effectively demonstrate the precise and controlled generation capabilities of our method. It is worth stressing that the end-to-end pipeline, including the graph-based knowledge representation, LLM-driven generation, and simulation of scenarios within BluebirdDT, delivers a complete and fully functional system for controllable air traffic scenario generation.

Furthermore, we observed that advanced reasoning models can respond to corrective feedback, refining a scenario based on its evaluated outcome. In particular, when asked to generate \textit{non-interacting} scenarios we observe that after providing feedback with specific information about aircraft pairs which violate the requirement, several models were able to adapt their scenarios to correct mistakes. For example, in Figure~\ref{fig:bad_example} we show one of the $N=30$ traffic volume benchmarks that Gemini-2.5-Pro fails to solve. Given the feedback on \textit{how it failed}, the model was able to adapt its solution and pass the benchmark. For verifiable scenario specifications (e.g., the specific number of aircraft or interactions), this feedback mechanism is automatic and scalable. The prompt used for the experiments in this section is detailed in the Supplementary Material.

\section{Discussion}
This paper introduces \texttt{AirTrafficGen}, a novel, end-to-end framework for configurable generation of air traffic scenarios, leveraging advanced reasoning capabilities of large language models. It offers a systematic, scalable alternative to the labour-intensive process of handcrafting scenarios. By encoding the three-dimensional problem into a graph-based representation that LLMs can process, we achieve fine-grained control over the characteristics of scenarios.

Our benchmarking methodology provides granular insights into LLM performance across crucial spatial, temporal and interaction reasoning axes. Our graph representation ensures that every generated scenario can be automatically rolled out and verified by counting interactions. State-of-the-art models such as Gemini-2.5-Pro and OpenAI's o3, GPT-oss and GPT-5 models reliably create high-traffic non-interacting scenarios and precisely engineer diverse interaction types, including pairwise conflicts and more complex multi-aircraft interactions. In the future, given the automatically verifiable nature of our benchmarks it would be interesting to study the effect of recent methods in prompt optimisation \cite{agrawal2025gepareflectivepromptevolution}.

A key strength of our method lies in its ability to \textbf{adapt existing scenarios}, tailoring them to specific instructions by adding new aircraft or modifying existing scenario data. This flexibility, along with sector-agnostic deployment, supports rapid prototyping and the creation of varied, challenging scenarios. Section~\ref{sec:controllability_section} demonstrated fine-grained controllability in a range of examples. Future work will include quantitative human-in-the-loop trials in which human ATCOs assess how well generated scenarios meet more nuanced specifications.

Looking ahead, future work will explore the integration of more complex operational elements, such as \textbf{holding patterns}, standard airport terminal departure and arrival procedures (SIDs and STARs), and the dynamic introduction of \textbf{unexpected events}. Investigating mechanisms for human feedback and editing within the generation loop could provide valuable refinements, combining an LLM's generative power with expert domain knowledge.

In conclusion, this research demonstrates the remarkable potential of LLMs in complex planning tasks within air traffic control. The framework overcomes the limits of traditional scenario generation and creates an opportunity for more efficient, adaptable, and diverse simulations for ATCO training and operational validation.

\bibliographystyle{plainnat}
\bibliography{airtrafficgen}

\appendix
\section{Example synthetic sector for benchmarking experiments}
\begin{figure}
    \centering
    \includegraphics[width=0.6\linewidth]{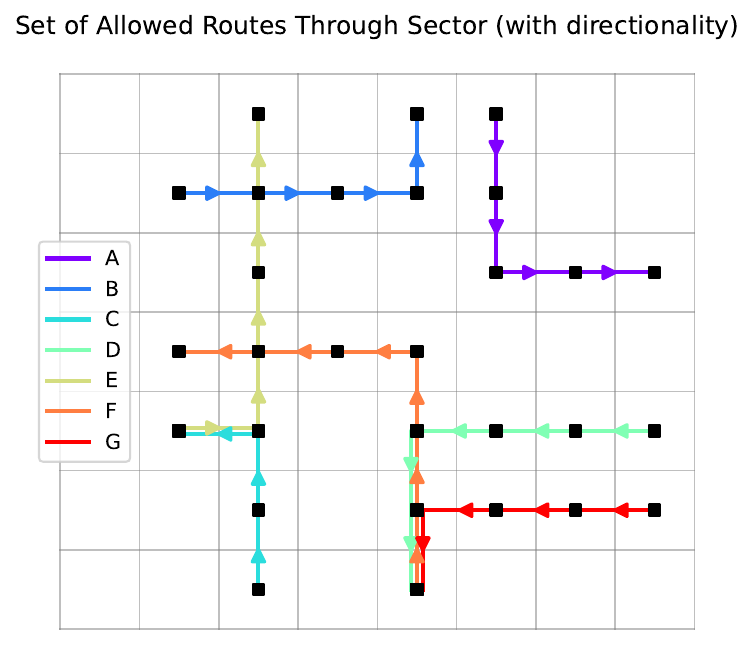}
    \caption{One of the synthetic sectors used in the benchmarking.}
    \label{fig:synthetic_sector}
\end{figure}

Figure~\ref{fig:synthetic_sector} illustrates the route structures in one of the synthetic sectors used for benchmarking. 
\FloatBarrier

\section{Full List of Models}
We provide a full list of models included in our benchmarking experiments in table \ref{tab:models_benchmarked}.

\begin{table}[h]
\centering
\caption{The models used in the benchmarking experiments and whether they support extended reasoning or thinking modes. Where not stated, model versions used were the latest available models on OpenRouter as of 12$^{\text{th}}$ August 2025. Pricing data reflects the \textit{average} cost per million output tokens on the OpenRouter platform as of 12$^{\text{th}}$ August 2025.}
\small 
\resizebox{\columnwidth}{!}{%
\begin{tabular}{|c|c|c|}
    \hline
    \textbf{Model} & \textbf{Reasoning} & \textbf{$\$$/M Output Tokens} \\
    \hline
    qwen-3-8b-it \cite{qwen3technicalreport} & $\checkmark$  & 0.138 \\
    mistral-small-3.2-24b-instruct-2506 \cite{mistralai2025mistralsmall24Binstruct} & -- &  0.20 \\
    llama-3.3-70b-it \cite{llama3.370b} & -- & 0.23  \\
    gpt-oss-20b \cite{openai_gpt-oss_2025} & \checkmark & 0.28 \\
    gemma-3-12b-it \cite{gemma_2025} & -- & 0.33 \\
    gemini-2.5-flash-lite-preview-06-17 \cite{deepmind2025gemini25flashlite} & $\checkmark$ & 0.40 \\
    gpt-oss-120b \cite{openai_gpt-oss_2025} & \checkmark & 0.48 \\
    llama-3.1-405b-it \cite{meta2024llama3_1_405b_instruct_hf} & -- & 2.51 \\
    deepseek-r1-0528 \cite{deepseekai2025deepseekr1incentivizingreasoningcapability} & \checkmark & 2.55 \\
    kimi-k2 \cite{moonshotai2025kimi_k2_blog} & -- & 2.58\\
    gpt-4.1-2025-04-14\cite{openai2025gpt4.1} & -- & 8.0 \\
    o3 \cite{openai2025o3} & $\checkmark$ & 8.0 \\
    gpt-5 \cite{openai_gpt5_2025} & \checkmark & 10.0 \\
    gemini-2.5-pro \cite{deepmind2025gemini25pro} & $\checkmark$ & 12.50 \\
\hline
\end{tabular}
}
\label{tab:models_benchmarked}
\end{table}

\section{Benchmark Results}
\label{sec:benchmark_results}
In Tables \ref{tab:num_ac_benchmark_table}, \ref{tab:scenario_len_benchmark_table}, \ref{tab:sector_complexity_table} and \ref{tab:input_conflict_table} we present the benchmarking results for the varying scenario traffic volume, length, sector complexity, and interaction number benchmarks, respectively.

\begin{table*}[htbp]
\small
\centering
\caption{Benchmarking Results for varying number of aircraft. Values quoted are the average number of unique interacting pairs of aircraft computed across 10 generated scenarios on 10 synthetic sectors. Optimal behaviour (marked with a checkmark) is zero interacting pairs.}
\label{tab:num_ac_benchmark_table}
\begin{tabular}{l|ccccccccccccc}
\toprule
Number of Aircraft & 2 & 3 & 4 & 5 & 6 & 7 & 8 & 9 & 10 & 15 & 20 & 25 & 30 \\
\midrule
random & 0.1 & 0.2 & 0.4 & 0.6 & 0.9 & 1.4 & 1.8 & 2.3 & 2.9 & 6.7 & 12.2 & 19.3 & 28.4 \\
\midrule
kimi-k2 & \cellcolor[RGB]{179,255,179}\checkmark & 0.1 & \cellcolor[RGB]{255,179,179}0.2 & 0.2 & 0.3 & 1.1 & 0.8 & 1.5 & 0.8 & 4.4 & \cellcolor[RGB]{255,179,179}9.6 & 11.1 & \cellcolor[RGB]{255,179,179}20.8 \\
gpt-4.1-2025-04-14 & \cellcolor[RGB]{179,255,179}\checkmark & \cellcolor[RGB]{179,255,179}\checkmark & \cellcolor[RGB]{179,255,179}\checkmark & 0.1 & 0.1 & 0.2 & \cellcolor[RGB]{179,255,179}\checkmark & 0.3 & 0.4 & 2.7 & 3.6 & 5.7 & 16.0 \\
gemini-2.5-flash-lite-preview-06-17 & \cellcolor[RGB]{179,255,179}\checkmark & 0.1 & \cellcolor[RGB]{179,255,179}\checkmark & 0.2 & 0.2 & 0.7 & 0.3 & 0.6 & 1.1 & 2.2 & 5.6 & 6.6 & 14.6 \\
qwen3-8b-it & \cellcolor[RGB]{179,255,179}\checkmark & 0.1 & 0.1 & \cellcolor[RGB]{255,179,179}0.6 & \cellcolor[RGB]{255,179,179}0.6 & 0.9 & \cellcolor[RGB]{255,179,179}2.0 & \cellcolor[RGB]{255,179,179}2.0 & \cellcolor[RGB]{255,179,179}1.2 & \cellcolor[RGB]{255,179,179}6.2 & 8.4 & \cellcolor[RGB]{255,179,179}17.3 & 12.2 \\
llama-3.3-70b-it & \cellcolor[RGB]{179,255,179}\checkmark & \cellcolor[RGB]{179,255,179}\checkmark & \cellcolor[RGB]{179,255,179}\checkmark &0.1 & 0.3 & 1.1 & 1.3 & 1.0 & 1.1 & 2.9 & 4.7 & 2.9 & 11.2 \\
llama-3.1-405b-it & \cellcolor[RGB]{179,255,179}\checkmark & 0.1 & 0.1 & 0.2 & 0.4 & \cellcolor[RGB]{255,179,179}1.2 & 0.9 & 1.5 & 1.5 & 3.5 & 8.0 & 6.6 & 11.0 \\
deepseek-r1-0528 & \cellcolor[RGB]{179,255,179}\checkmark & \cellcolor[RGB]{179,255,179}\checkmark & \cellcolor[RGB]{179,255,179}\checkmark & \cellcolor[RGB]{179,255,179}\checkmark & \cellcolor[RGB]{179,255,179}\checkmark & 0.1 & 0.2 & \cellcolor[RGB]{179,255,179}\checkmark & 0.4 & 1.3 & 4.0 & 6.1 & 9.2 \\
gemma-3-12b-it & \cellcolor[RGB]{179,255,179}\checkmark & \cellcolor[RGB]{179,255,179}\checkmark & 0.1 & \cellcolor[RGB]{179,255,179}0.1 & 0.3 & 0.4 & 0.8 & 0.9 & 1.5 & 5.4 & 3.3 & 8.8 & 9.0 \\
mistral-small-3.2-24b-instruct-2506 & \cellcolor[RGB]{179,255,179}\checkmark & \cellcolor[RGB]{179,255,179}\checkmark & \cellcolor[RGB]{255,179,179}0.2 & \cellcolor[RGB]{179,255,179}\checkmark & 0.5 & 1.1 & 0.7 & 1.1 & 1.3 & 4.2 & 7.0 & 9.9 & 5.4 \\
gemini-2.5-pro & \cellcolor[RGB]{179,255,179}\checkmark & \cellcolor[RGB]{179,255,179}\checkmark & \cellcolor[RGB]{179,255,179}\checkmark & \cellcolor[RGB]{179,255,179}\checkmark & \cellcolor[RGB]{179,255,179}\checkmark & \cellcolor[RGB]{179,255,179}\checkmark & \cellcolor[RGB]{179,255,179}\checkmark & \cellcolor[RGB]{179,255,179}\checkmark & 0.1 & 0.6 & 0.3 & 0.3 & 0.9 \\
gpt-oss-20b & \cellcolor[RGB]{179,255,179} \checkmark & \cellcolor[RGB]{179,255,179} \checkmark & \cellcolor[RGB]{179,255,179} \checkmark & \cellcolor[RGB]{179,255,179} \checkmark & \cellcolor[RGB]{179,255,179} \checkmark& \cellcolor[RGB]{179,255,179} \checkmark& \cellcolor[RGB]{179,255,179} \checkmark & \cellcolor[RGB]{179,255,179} \checkmark & 0.1 & 0.1 & 0.7 & 1.0 & 0.2 \\
o3 & \cellcolor[RGB]{179,255,179}\checkmark & \cellcolor[RGB]{179,255,179}\checkmark & \cellcolor[RGB]{179,255,179}\checkmark & \cellcolor[RGB]{179,255,179}\checkmark & \cellcolor[RGB]{179,255,179}\checkmark & \cellcolor[RGB]{179,255,179}\checkmark & \cellcolor[RGB]{179,255,179}\checkmark & \cellcolor[RGB]{179,255,179}\checkmark & \cellcolor[RGB]{179,255,179}\checkmark & \cellcolor[RGB]{179,255,179}\checkmark & 0.2 & 0.2 & \cellcolor[RGB]{179,255,179}\checkmark \\
gpt-oss-120b & \cellcolor[RGB]{179,255,179} \checkmark & \cellcolor[RGB]{179,255,179} \checkmark & \cellcolor[RGB]{179,255,179} \checkmark & \cellcolor[RGB]{179,255,179} \checkmark & \cellcolor[RGB]{179,255,179} \checkmark & 0.1 & 0.1 & \cellcolor[RGB]{179,255,179} \checkmark & 0.2 & \cellcolor[RGB]{179,255,179} \checkmark & 0.1 & \cellcolor[RGB]{179,255,179} \checkmark & \cellcolor[RGB]{179,255,179} \checkmark \\
gpt-5-low & \cellcolor[RGB]{179,255,179} \checkmark & \cellcolor[RGB]{179,255,179} \checkmark & \cellcolor[RGB]{179,255,179} \checkmark & \cellcolor[RGB]{179,255,179} \checkmark & \cellcolor[RGB]{179,255,179} \checkmark & \cellcolor[RGB]{179,255,179}\checkmark & \cellcolor[RGB]{179,255,179}\checkmark & \cellcolor[RGB]{179,255,179} \checkmark & \cellcolor[RGB]{179,255,179}\checkmark & \cellcolor[RGB]{179,255,179} \checkmark & \cellcolor[RGB]{179,255,179}\checkmark & \cellcolor[RGB]{179,255,179} \checkmark & \cellcolor[RGB]{179,255,179} \checkmark \\
gpt-5-high & \cellcolor[RGB]{179,255,179} \checkmark & \cellcolor[RGB]{179,255,179} \checkmark & \cellcolor[RGB]{179,255,179} \checkmark & \cellcolor[RGB]{179,255,179} \checkmark & \cellcolor[RGB]{179,255,179} \checkmark & \cellcolor[RGB]{179,255,179}\checkmark & \cellcolor[RGB]{179,255,179}\checkmark & \cellcolor[RGB]{179,255,179} \checkmark & \cellcolor[RGB]{179,255,179}\checkmark & \cellcolor[RGB]{179,255,179} \checkmark & \cellcolor[RGB]{179,255,179}\checkmark & \cellcolor[RGB]{179,255,179} \checkmark & \cellcolor[RGB]{179,255,179}\checkmark \\

\bottomrule
\end{tabular}
\end{table*}

\begin{table*}[htbp]
\small
\centering
\caption{Benchmarking results for varying scenario length. Values quoted are the average number of unique interacting pairs of aircraft computed across 10 generated scenarios on 10 synthetic sectors. Optimal behaviour (marked with a checkmark) is zero interacting pairs.}
\label{tab:scenario_len_benchmark_table}
\begin{tabular}{l|ccccc}
\toprule
Scenario Length & 12 & 15 & 18 & 21 & 24 \\
\midrule
random baseline & 1.8 & 1.7 & 1.6 & 1.4 & 1.2 \\
\midrule
qwen3-8b-it & \cellcolor[RGB]{255,179,179}1.3 & 1.2 & \cellcolor[RGB]{255,179,179}1.0 & \cellcolor[RGB]{255,179,179}1.7 & \cellcolor[RGB]{255,179,179}1.2 \\
kimi-k2 & 1.0 & \cellcolor[RGB]{255,179,179}1.6 & \cellcolor[RGB]{255,179,179}1.0 & 0.4 & 0.9 \\
llama-3.1-405b-it & \cellcolor[RGB]{255,179,179}1.3 & 0.9 & 1.3 & 0.7 & 0.9 \\
gemma-3-12b-it & 0.6 & 0.9 & 0.9 & 1.2 & 0.6 \\
mistral-small-3.2-24b-instruct-2506 & 0.6 & 1.0 & 0.7 & 1.0 & 0.5 \\
gemini-2.5-flash-lite-preview-06-17 & 0.4 & 0.4 & 0.3 & 0.2 & 0.2 \\
gpt-4.1-2025-04-14 & \cellcolor[RGB]{179,255,179}\checkmark & 0.4 & 0.1 & 0.1 & 0.1 \\
llama-3.3-70b-it & 0.8 & 0.4 & 0.9 & 0.3 & 0.1 \\
gemini-2.5-pro & \cellcolor[RGB]{179,255,179}\checkmark & \cellcolor[RGB]{179,255,179}\checkmark & \cellcolor[RGB]{179,255,179}\checkmark & \cellcolor[RGB]{179,255,179}\checkmark & 0.1 \\
deepseek-r1-0528 & 0.2 & \cellcolor[RGB]{179,255,179}\checkmark & 0.1 & \cellcolor[RGB]{179,255,179}\checkmark & \cellcolor[RGB]{179,255,179}\checkmark \\
gpt-oss-20b & 0.1 & \cellcolor[RGB]{179,255,179}\checkmark & \cellcolor[RGB]{179,255,179}\checkmark & 0.1 & \cellcolor[RGB]{179,255,179}\checkmark \\
o3 & \cellcolor[RGB]{179,255,179}\checkmark & \cellcolor[RGB]{179,255,179}\checkmark & \cellcolor[RGB]{179,255,179}\checkmark & \cellcolor[RGB]{179,255,179}\checkmark & \cellcolor[RGB]{179,255,179}\checkmark \\
gpt-oss-120b & \cellcolor[RGB]{179,255,179}\checkmark & \cellcolor[RGB]{179,255,179}\checkmark & \cellcolor[RGB]{179,255,179}\checkmark & \cellcolor[RGB]{179,255,179}\checkmark & \cellcolor[RGB]{179,255,179}\checkmark \\
gpt-5-low & \cellcolor[RGB]{179,255,179}\checkmark & \cellcolor[RGB]{179,255,179}\checkmark & \cellcolor[RGB]{179,255,179}\checkmark & \cellcolor[RGB]{179,255,179}\checkmark & \cellcolor[RGB]{179,255,179}\checkmark \\
gpt-5-high& \cellcolor[RGB]{179,255,179}\checkmark & \cellcolor[RGB]{179,255,179}\checkmark & \cellcolor[RGB]{179,255,179}\checkmark & \cellcolor[RGB]{179,255,179}\checkmark & \cellcolor[RGB]{179,255,179}\checkmark \\

\bottomrule
\end{tabular}
\end{table*}

\begin{table*}[htbp]
\small
\centering
\caption{Benchmarking results for increasing sector complexity. Values quoted are the average number of unique interacting pairs of aircraft computed across 10 generated scenarios on 10 synthetic sectors. Optimal behaviour (marked with a checkmark) is zero interacting pairs.}
\label{tab:sector_complexity_table}
\begin{tabular}{l|ccccccccccc}
\toprule
Sector Route Intersections & 4 & 5 & 6 & 7 & 8 & 9 & 10 & 11 & 12 & 13 & 14 \\
\midrule
random & 1.9 & 1.9 & 1.9 & 2.3 & 2.3 & 2.7 & 2.5 & 2.7 & 2.9 & 2.7 & 3.1 \\
\midrule
qwen3-8b-it & \cellcolor[RGB]{255,179,179}1.3 & \cellcolor[RGB]{255,179,179}1.6 & \cellcolor[RGB]{255,179,179}1.2 & \cellcolor[RGB]{255,179,179}2.3 & 1.4 & \cellcolor[RGB]{255,179,179}2.4 & \cellcolor[RGB]{255,179,179}2.3 & \cellcolor[RGB]{255,179,179}2.6 & \cellcolor[RGB]{255,179,179}3.8 & \cellcolor[RGB]{255,179,179}3.5 & \cellcolor[RGB]{255,179,179}3.1 \\
kimi-k2 & 1.2 & \cellcolor[RGB]{255,179,179}1.6 & 0.9 & 1.4 & \cellcolor[RGB]{255,179,179}1.7 & 2.0 & 2.0 & 1.4 & 2.1 & 2.2 & 3.0 \\
mistral-small-3.2-24b-instruct-2506 & 1.0 & 0.7 & 1.1 & 1.4 & 1.9 & 1.8 & 1.9 & 1.9 & 2.3 & 1.8 & 2.9 \\
gemma-3-12b-it & 1.2 & 1.2 & 0.8 & 1.7 & 0.9 & 1.5 & 1.1 & 2.1 & 2.3 & 2.8 & 2.8 \\
llama-3.1-405b-it & 0.5 & 1.2 & 0.6 & 1.0 & 0.9 & 1.9 & 1.8 & 1.8 & 1.6 & 1.9 & 2.4 \\
deepseek-r1-0528 & \cellcolor[RGB]{179,255,179}\checkmark & \cellcolor[RGB]{179,255,179}\checkmark & 0.1 & 0.6 & 0.8 & 0.7 & 0.2 & 0.4 & 0.8 & 0.5 & 1.7 \\
gemini-2.5-flash-lite-preview-06-17 & 0.6 & 0.9 & 0.4 & 0.2 & 0.4 & 0.9 & 0.9 & 1.2 & 1.0 & 1.3 & 1.5 \\
llama-3.3-70b-it & 1.0 & 0.5 & 0.4 & 1.0 & 1.2 & 0.8 & 1.2 & 0.7 & 1.7 & 1.5 & 1.2 \\
gpt-4.1-2025-04-14 & 0.3 & 0.1 & 0.2 & 0.5 & 0.4 & 0.7 & 0.2 & 0.2 & 0.9 & 0.7 & 0.8 \\
gemini-2.5-pro & \cellcolor[RGB]{179,255,179}\checkmark & \cellcolor[RGB]{179,255,179}\checkmark & \cellcolor[RGB]{179,255,179}\checkmark & \cellcolor[RGB]{179,255,179}\checkmark & \cellcolor[RGB]{179,255,179}\checkmark & \cellcolor[RGB]{179,255,179}\checkmark & \cellcolor[RGB]{179,255,179}\checkmark & 0.1 & 0.5 & 0.1 & \cellcolor[RGB]{179,255,179}\checkmark \\
gpt-oss-20b & \cellcolor[RGB]{179,255,179}\checkmark  & \cellcolor[RGB]{179,255,179}\checkmark  & \cellcolor[RGB]{179,255,179}\checkmark  & 0.1 & 0.1 & 0.1 & 0.1 & \cellcolor[RGB]{179,255,179}\checkmark  & 0.1 & \cellcolor[RGB]{179,255,179}\checkmark  & \cellcolor[RGB]{179,255,179}\checkmark  \\

o3 & \cellcolor[RGB]{179,255,179}\checkmark & \cellcolor[RGB]{179,255,179}\checkmark & \cellcolor[RGB]{179,255,179}\checkmark & \cellcolor[RGB]{179,255,179}\checkmark & \cellcolor[RGB]{179,255,179}\checkmark & \cellcolor[RGB]{179,255,179}\checkmark & \cellcolor[RGB]{179,255,179}\checkmark & \cellcolor[RGB]{179,255,179}\checkmark & \cellcolor[RGB]{179,255,179}\checkmark & \cellcolor[RGB]{179,255,179}\checkmark & \cellcolor[RGB]{179,255,179}\checkmark \\
gpt-oss-120b & \cellcolor[RGB]{179,255,179}\checkmark  & \cellcolor[RGB]{179,255,179}\checkmark  & \cellcolor[RGB]{179,255,179}\checkmark  & 0.2 & \cellcolor[RGB]{179,255,179}\checkmark  & 0.1 & \cellcolor[RGB]{179,255,179}\checkmark  & \cellcolor[RGB]{179,255,179}\checkmark  & 0.2 & 0.3 & \cellcolor[RGB]{179,255,179}\checkmark  \\
gpt-5-low & \cellcolor[RGB]{179,255,179}\checkmark& \cellcolor[RGB]{179,255,179}\checkmark & \cellcolor[RGB]{179,255,179}\checkmark & \cellcolor[RGB]{179,255,179}\checkmark & \cellcolor[RGB]{179,255,179}\checkmark & \cellcolor[RGB]{179,255,179}\checkmark & \cellcolor[RGB]{179,255,179}\checkmark & \cellcolor[RGB]{179,255,179}\checkmark & 0.1 & \cellcolor[RGB]{179,255,179}\checkmark & \cellcolor[RGB]{179,255,179}\checkmark \\
gpt-5-high & \cellcolor[RGB]{179,255,179}\checkmark & \cellcolor[RGB]{179,255,179}\checkmark & \cellcolor[RGB]{179,255,179}\checkmark & \cellcolor[RGB]{179,255,179}\checkmark & \cellcolor[RGB]{179,255,179}\checkmark & \cellcolor[RGB]{179,255,179}\checkmark & \cellcolor[RGB]{179,255,179}\checkmark & \cellcolor[RGB]{179,255,179}\checkmark & \cellcolor[RGB]{179,255,179}\checkmark & 0.1 & \cellcolor[RGB]{179,255,179}\checkmark \\
\bottomrule
\end{tabular}
\end{table*}

\begin{table*}[htbp]
\small
\centering
\caption{Benchmarking results for increasing number of input conflicts. Values quoted are the mean absolute difference between the number of generated interacting pairs of aircraft and the input number. This average is computed across 10 generated scenarios on 10 synthetic sectors. Checkmarks denote perfect performance across all 10 synthetic sectors (a mean absolute difference of zero).}
\label{tab:input_conflict_table}
\begin{tabular}{l|ccccc}
\toprule
Number of Interactions & 1.0 & 2.0 & 3.0 & 4.0 & 5.0 \\
\midrule
random baseline & 1.95 & 1.55 & 1.49 & 1.92 & 2.58 \\
\midrule
llama-3.3-70b-it & 1.30 & 1.30 & 1.70 & 2.20 & \cellcolor[RGB]{255,179,179}4.20 \\
deepseek-r1-0528 & 0.30 & 0.60 & 0.90 & 1.20 & 3.90 \\
openai/gpt-oss-20b & \cellcolor[RGB]{179,255,179}\checkmark & \cellcolor[RGB]{179,255,179}\checkmark & 1.10 & \cellcolor[RGB]{255,179,179}3.00 & 3.80 \\
gpt-4.1-2025-04-14 & 0.50 & \cellcolor[RGB]{255,179,179}2.20 & 1.40 & 2.40 & 3.40 \\
gemini-2.5-flash-lite-preview-06-17 & \cellcolor[RGB]{255,179,179}1.90 & 2.00 & \cellcolor[RGB]{255,179,179}3.30 & 1.20 & 3.00 \\
gemma-3-12b-it & 1.20 & 1.20 & 1.60 & 2.20 & 2.90 \\
llama-3.1-405b-it & 0.90 & 1.50 & 2.20 & 2.10 & 2.80 \\
qwen3-8b-it & 1.80 & 1.90 & 2.10 & 2.20 & 2.80 \\
mistral-small-3.2-24b-instruct-2506 & 0.90 & 1.50 & 1.50 & 1.90 & 2.50 \\
kimi-k2 & 0.90 & 1.20 & 1.30 & 2.20 & 1.90 \\
gemini-2.5-pro & \cellcolor[RGB]{179,255,179}\checkmark & 0.10 & 0.60 & 1.50 & 1.70 \\
openai/gpt-oss-120b & \cellcolor[RGB]{179,255,179}\checkmark & 0.50 & 0.50 & 1.60 & 0.80 \\
o3 & \cellcolor[RGB]{179,255,179}\checkmark & \cellcolor[RGB]{179,255,179}\checkmark & \cellcolor[RGB]{179,255,179}\checkmark & 0.40 & 0.30 \\
gpt-5-high & \cellcolor[RGB]{179,255,179}\checkmark & \cellcolor[RGB]{179,255,179}\checkmark & 0.10 & 0.20 & 0.20 \\
gpt-5-low & \cellcolor[RGB]{179,255,179}\checkmark & \cellcolor[RGB]{179,255,179}\checkmark & 0.10 & \cellcolor[RGB]{179,255,179}\checkmark & \cellcolor[RGB]{179,255,179}0.10 \\
\bottomrule
\end{tabular}
\end{table*}

\begin{figure}
    \centering
    \includegraphics[width=\linewidth, trim= 0 13cm 0 0, clip]{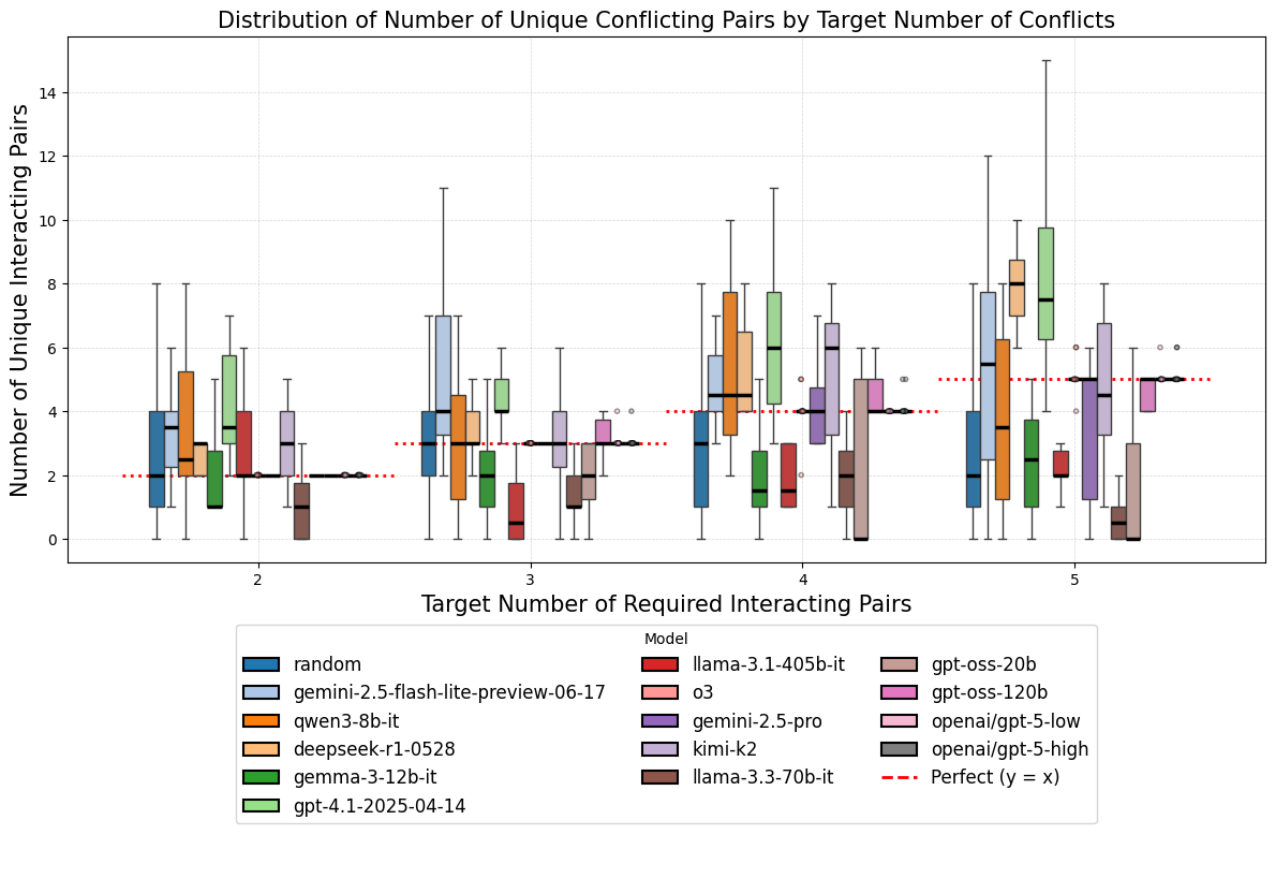}
    \caption{Benchmarking results for increasing number of input conflicts. o3 and GPT-5 saturate the metric, and therefore we include their individual data points to maintain readability.}
    \label{fig:num_conflict_box_plot}
\end{figure}

Figure~\ref{fig:num_conflict_box_plot} summarises the controllability benchmark: box plots of the number of interactions each model generates at different targets, revealing systematic under- and over-generation. While o3 stays close to the targets in all settings, Gemini-2.5-Pro struggles with higher targets. Most models increase conflicts as the target increases, but do not reliably match the requested counts, indicating limited fine-grained control.

\begin{figure}
    \centering
    \includegraphics[width=\linewidth]{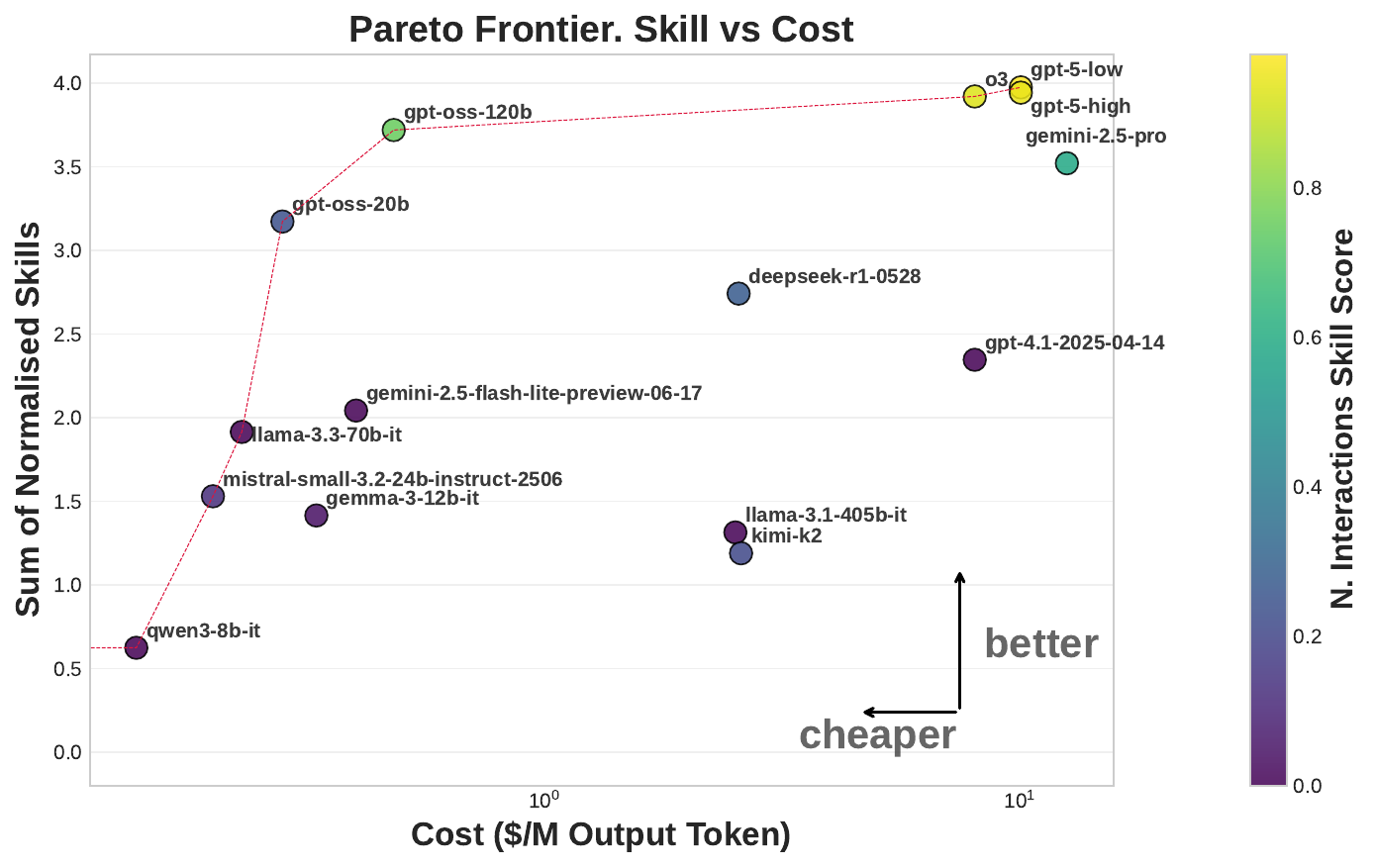}
    \caption{Comparing sum of normalised skills across all four benchmarks against inference cost for all models. The red dotted line represents the Pareto Fronter: lying on this line means that there is no model which is both cheaper and more skilful.}
    \label{fig:pareto_plot}
\end{figure}

Figure~\ref{fig:pareto_plot} compares the overall skill level across the four benchmarks (computed as the sum of the normalised skills) against inference cost. The dotted red line represents the Pareto front. It is particularly interesting to observe the GPT-oss models achieve close to optimal scores at a fraction of the cost.

\subsection{Case Study: High-Volume Scenario Generation ($N=30$)}
\begin{figure}
    \centering
        \includegraphics[width=\linewidth]{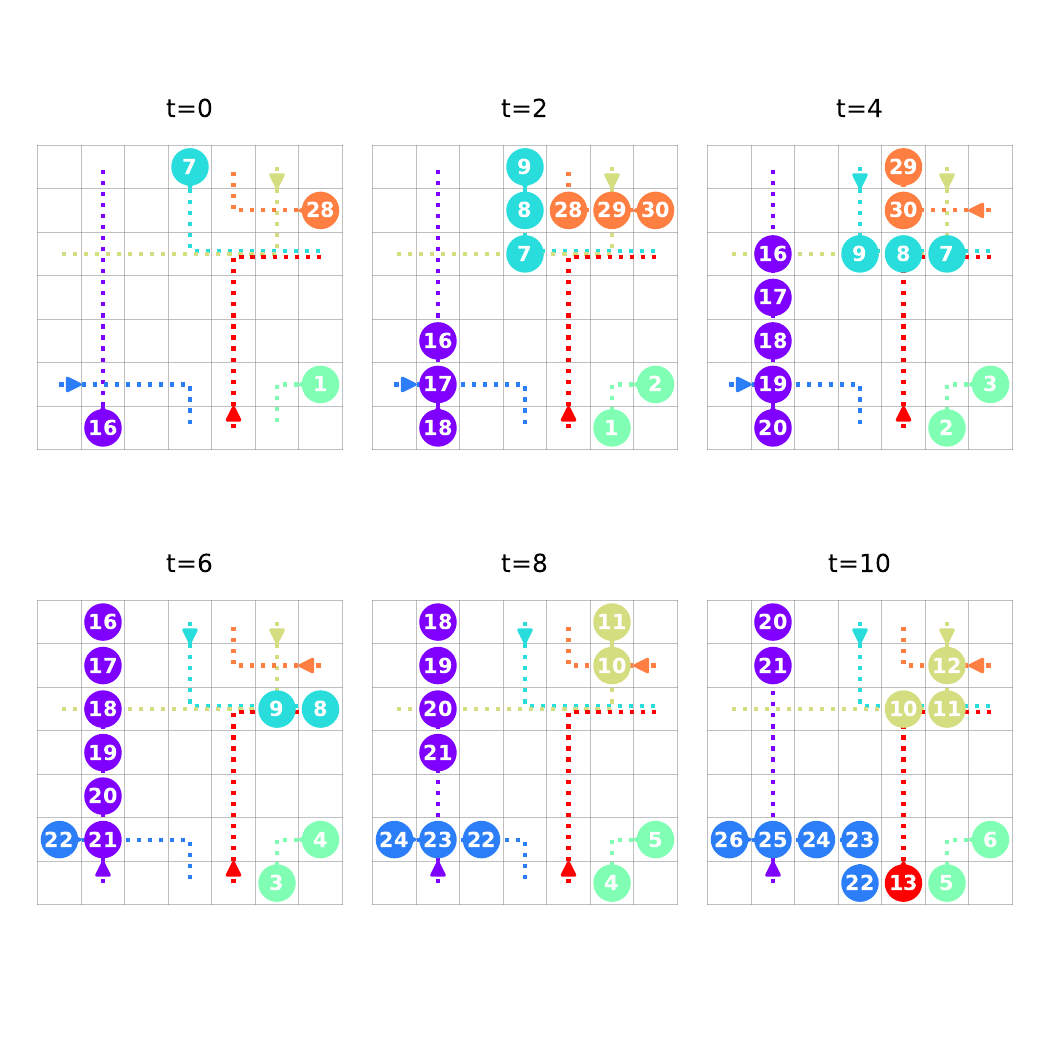}
    \caption{Gemini-2.5-Pro \textbf{solving} the $N=30$ benchmark: there are no interactions in this scenario. It devises a strategy where aircraft are systematically spawned in sequences along the same route, thus avoiding conflicts. Sequences of aircraft on intersecting routes avoid one another by careful selection of spawn times. (Snapshots shown every two time units for convenience).}
    \label{fig:good_example}
\end{figure}

\begin{figure}
    \centering
    \includegraphics[width=\linewidth]{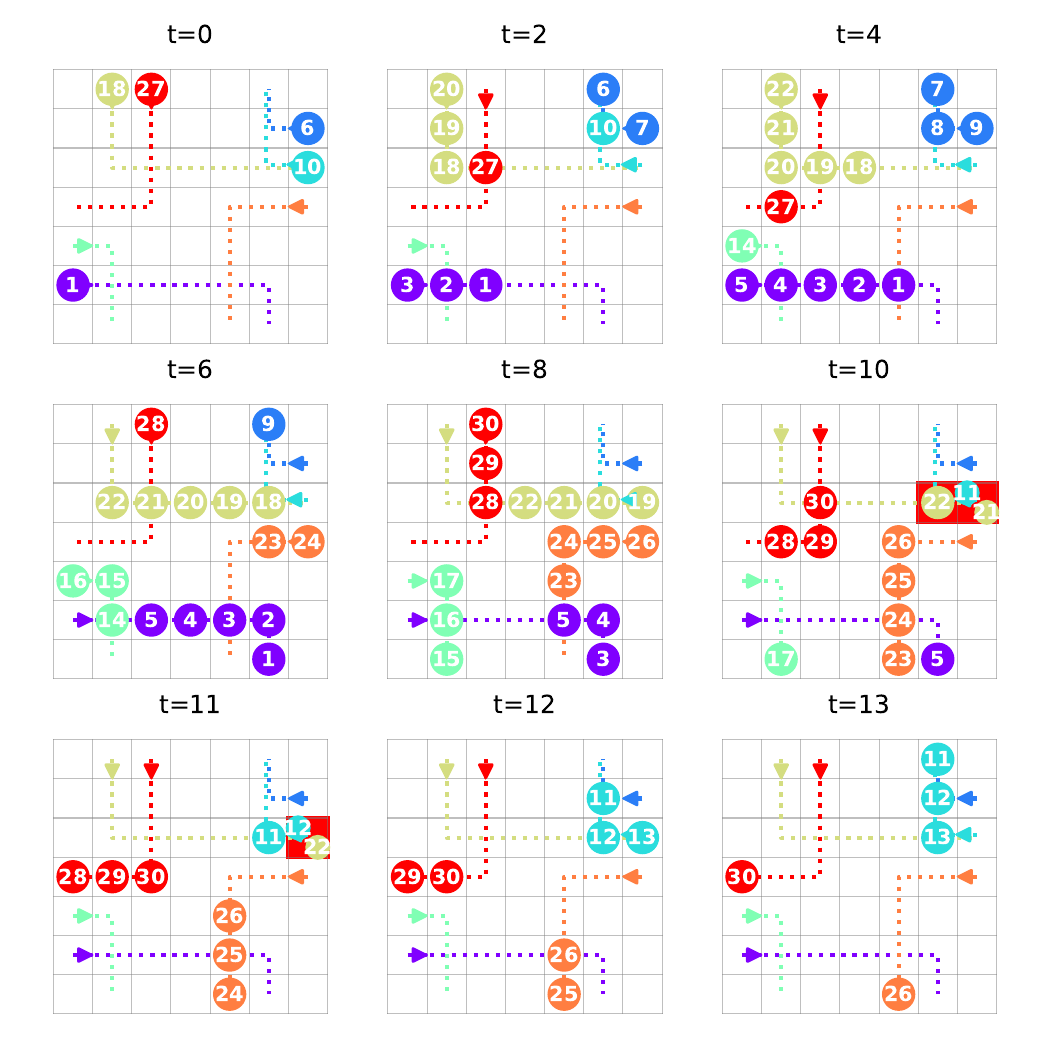}
    \caption{Gemini-2.5-Pro \textbf{failing to solve} an $N=30$ benchmark: there are three interactions in this scenario. In this case, at time $t=10,11$ aircraft 11 interacts with aircraft 21 and 22 respectively, in a head-on manner. Similarly, at $t=11$, aircraft 12 interacts with aircraft 22. Notice that, similar to Figure~\ref{fig:good_example}, the model has attempted to thread sequences of aircraft along the same route one-by-one. In this case, it has miscalculated the time at which aircraft 21 would reach the end of its route. (Snapshots shown every two time units for convenience).}
    \label{fig:bad_example}
\end{figure}

We present two Gemini-2.5-Pro outputs on the $N=30$ traffic-volume benchmark (Figures~\ref{fig:good_example}~and~\ref{fig:bad_example}). The model is successful in Figure~\ref{fig:good_example} and fails in Figure~\ref{fig:bad_example}, illustrating both the complexity of the task and, in the success case, the strategy employed by a reasoning-capable model.

\section{Extra Controllability Experiments}
\label{sec:controllability_appendix}
\begin{figure}[htbp]
    \centering
    \begin{subfigure}{0.32\textwidth} 
        \centering
        \includegraphics[width=\linewidth]{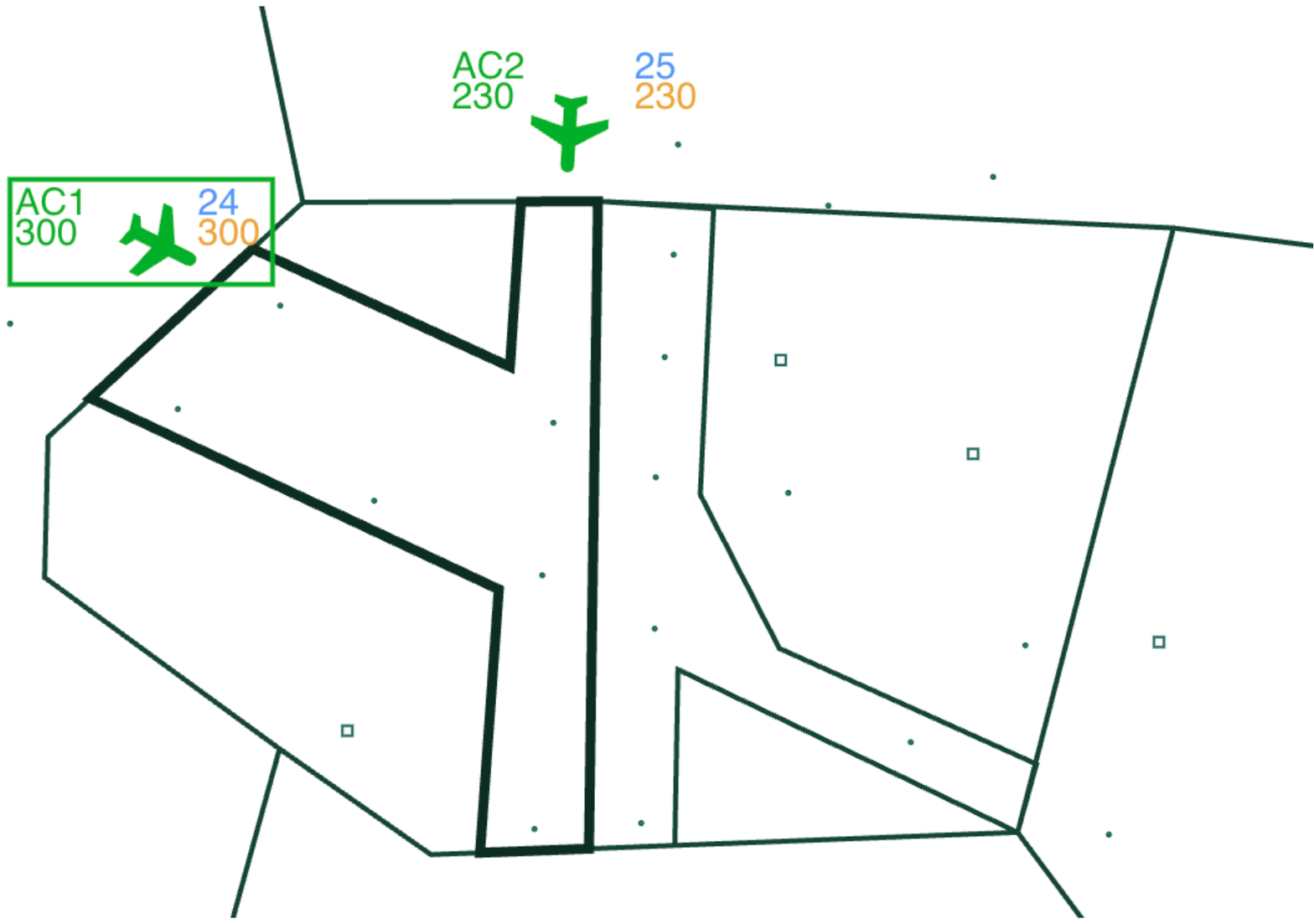}
        \label{fig:x-path}
    \caption{Generate a scenario in which two aircraft interact in a crossing-paths manner.}
    \end{subfigure}
    \hfill 
    \begin{subfigure}{0.32\textwidth} 
        \centering
        \includegraphics[width=\linewidth]{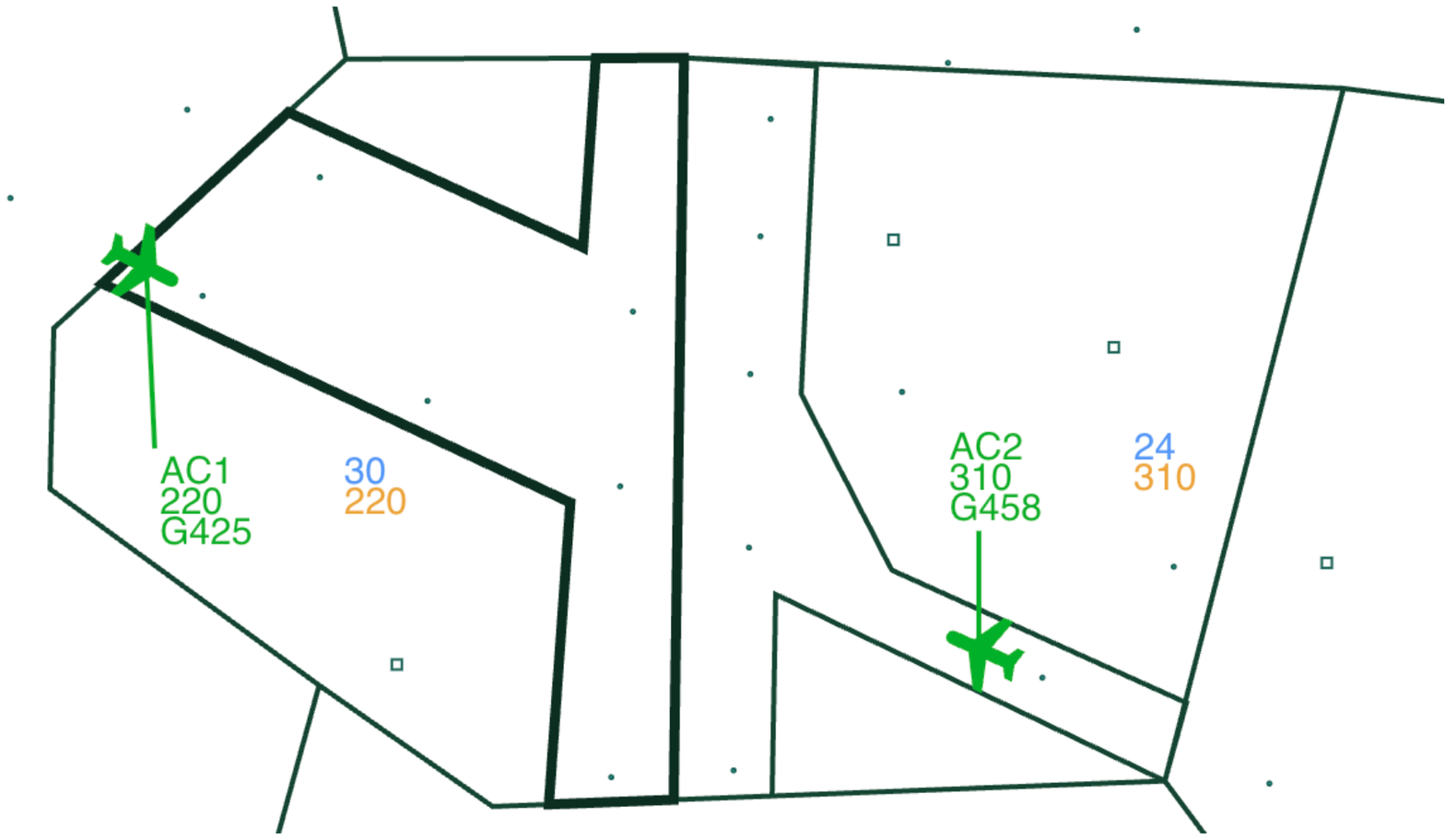}
        \label{fig:head-on}
    \caption{Generate a scenario in which two aircraft interact in a head-on manner.}
    \end{subfigure}
    \hfill 
    \begin{subfigure}{0.32\textwidth} 
        \centering
        \includegraphics[width=\linewidth]{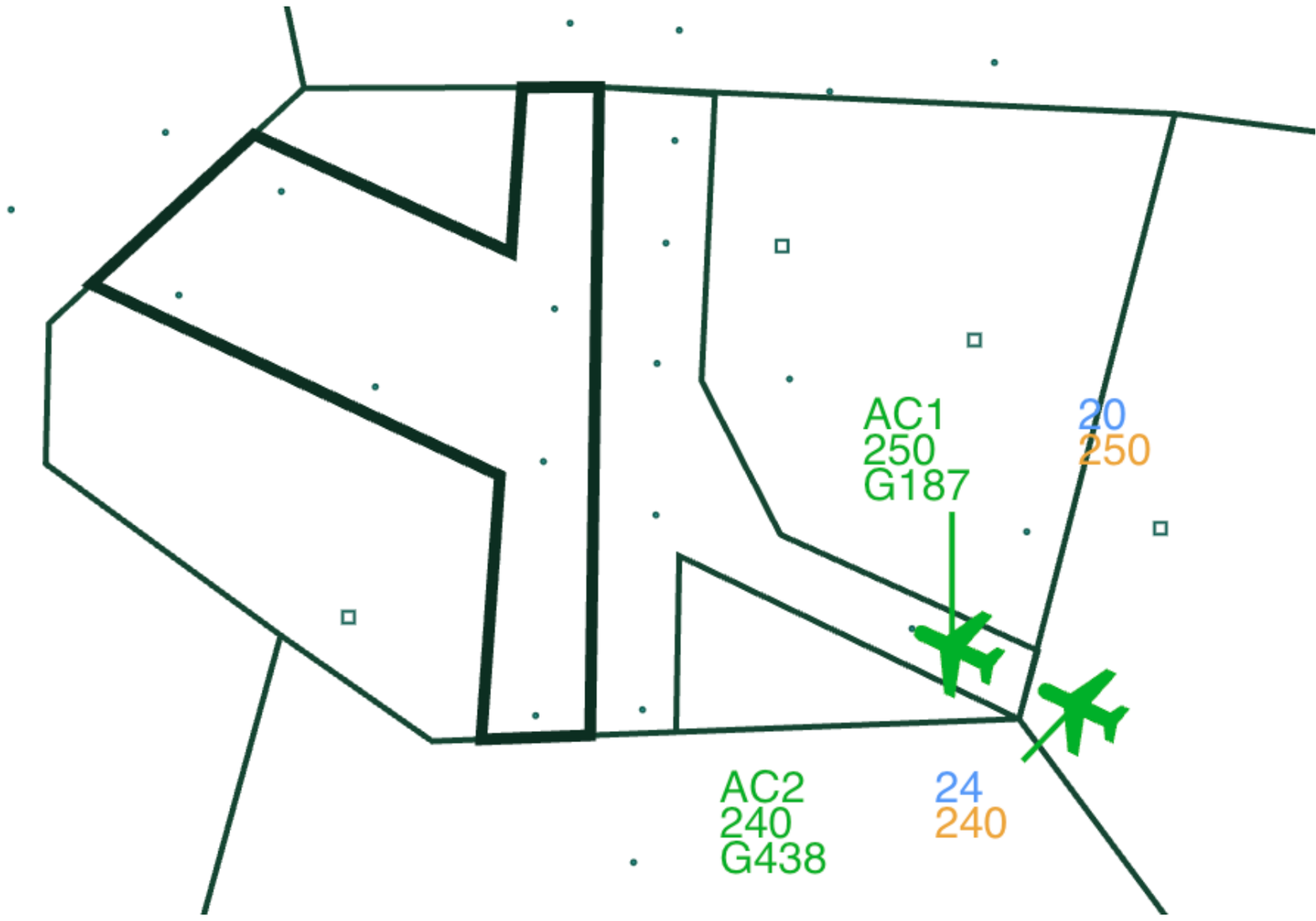}
        \label{fig:catch-up}
        \caption{Generate a scenario in which two aircraft interact in a catch-up configuration.}
    \end{subfigure}

    \caption{Three examples of fine-grained scenario controllability using AirTrafficGen. Note that in (c) the two aircraft have two very different ground speeds (denoted GXXX), meaning that AC2 will catch up and overtake AC1. Note that differing ground speeds arise from the simulator when converting ``slow" and ``fast" movers in our discretised scheme to turboprops and jets respectively. The same aircraft type will fly at different speeds according to its flight level.}
    \label{fig:pairwise_examples}
\end{figure}

In Figure~\ref{fig:pairwise_examples} we show that the method is able to accurately generate all three types of pairwise interactions.

\end{document}